\pdfoutput=1
\documentclass[11pt]{article} 
\usepackage{preamble} 
\usepackage{ACL2023}
\usepackage{csquotes}
\usepackage{enumitem}
\usepackage{metrix}
\usepackage{booktabs}
\usepackage{floatrow}
\usepackage{threeparttable}
\usepackage{multirow, makecell}

\graphicspath{{./plots/}}

\setlist[enumerate]{label= (\roman*)}

\renewfloatcommand{ttabbox}{table}[\nocapbeside][\FBwidth]

\title{\bygpt: End-to-End Style-conditioned Poetry Generation\\with Token-free Language Models}

\author{%
  Jonas Belouadi\\
  \mail{jonas.belouadi@uni-bielefeld.de}
  \And%
  Steffen Eger\\
  \mail{steffen.eger@uni-bielefeld.de}
  \AND%
  {\rm Natural Language Learning Group (NLLG)}\\
  Faculty of Technology, Bielefeld University\\
  \spurl{nl2g.github.io}
}

\begin{document}
\maketitle
\begin{abstract}
  State-of-the-art poetry generation systems are often complex. They either
  consist of task-specific model pipelines, incorporate prior knowledge in the
  form of manually created constraints, or both. In contrast, end-to-end models
  would not suffer from the overhead of having to model prior knowledge and
  could learn the nuances of poetry from data alone, reducing the degree of
  human supervision required. In this work, we investigate end-to-end poetry
  generation conditioned on styles such as rhyme, meter, and alliteration. We
  identify and address lack of training data and mismatching tokenization
  algorithms as possible limitations of past attempts. In particular, we
  successfully pre-train \bygpt, a new token-free decoder-only language model,
  and fine-tune it on a large custom corpus of English and German quatrains
  annotated with our styles. We show that \bygpt outperforms other models such
  as \tfive[m], \tfive[By], \gpt and \chatgpt, while also being more parameter
  efficient and performing favorably compared to humans. In addition, we
  analyze its runtime performance and demonstrate that it is not prone to
  memorization. We make our code, models, and datasets publicly
  available.\insert\footins{\pdfrunninglinkoff}\footnote{%
    \anonymize{\url{https://github.com/potamides/uniformers}}%
  }\insert\footins{\pdfrunninglinkon} 
\end{abstract}

\section{Introduction}\label{sec:introduction} 
End-to-end fine-tuning of pre-trained language models like
\gpt~\citep{Radford2019} or \tfive~\citep{Raffel2020} on downstream tasks has
been an immensely popular training paradigm for text-generation in the last few
years~\citep{ijcai2021p612}. End-to-end models learn to complete a task by
directly learning all steps, without intermediary algorithms such as
hand-crafted rules or post-processing. This approach has proven to be highly
effective on a wide range of problems such as dialog
generation~\citep{sun-etal-2022-bort,Yang_Li_Quan_2021},
summarization~\citep{zhu-etal-2021-mediasum,zhong-etal-2021-qmsum,huang-etal-2021-efficient},
and machine translation~\citep{farinha-EtAl:2022:WMT,Tran2020CrosslingualRF}.
Nevertheless, all these applications have in common that they only concern
themselves with the generation of prosaic texts. Generating \emph{formal verse
poetry} on the other hand, with strict constraints on \emph{aesthetic style}
such as rhyme scheme, meter and alliteration, remains a difficult problem.
Attempts to employ end-to-end solutions in this context have so far been
unsuccessful~\citep{wockener-etal-2021-end}, with some authors even concluding
that language models cannot pick up such constraints from data
alone~\citep{popescu-belis-etal-2022-constrained}.
As a consequence, state-of-the-art poetry generation systems rely on human
guidance by (i) injecting prior knowledge\footnote{We define incorporating
prior knowledge as \enquote{Any form of influence on model decisions not
learned by the model itself.}} in the form of hard-coded constraints to filter
model outputs or modify probability distributions or (ii) breaking the whole
process down into sophisticated task-specific model pipelines.

\begin{figure}[t]
  \centering
  \stanza[\bygpt]{ABBA}
    {\metrics{u _ u _ u _ u _}
      {The \allit{s}weet wild \allit{s}train, the \allit{s}ud-den \A{\allit{s}tart},}}
    {\metrics{u _ u _ u _ u _}
      {Which shakes the per-\allit{f}umed al-tar's \B{\allit{f}lame},}}
    {\metrics{u _ u _ u _ u _}
      {To make \allit{i}ts shrine \allit{a} sa-cred \B{name},}}
    {\metrics{u _ u _ u _ u _}
      {\allit{A}nd sing \allit{i}ts praise \allit{i}n \allit{e}v-ery \A{heart}.}}
  \caption{Generated quatrain with ABBA rhyme scheme, high amount of
  alliterations (green), and iambic meter, i.e., unstressed
  syllable~(\metricsymbols*{u}) follows stressed syllable~(\metricsymbols*{_}).}%
  \label{fig:example-quatrain}
\end{figure}

\citet{tian-peng-2022-zero}, for example, propose a sonnet generation framework
with four distinct pipeline steps: content planning, rhyme pairs generation,
polishing for aesthetics, and finally sketch-to-sonnet generation. Further,
they incorporate prior knowledge such as pronunciation dictionaries, knowledge
bases, and lexically constrained decoding. Similarly,
\citet{hopkins-kiela-2017-automatically} use Weighted Finite State Transducers
to monitor whether their poetry generation system meets metric constraints and
roll back its state in case of a violation.

Such forms of human supervision lead to ramifications that an end-to-end
solution would not face. Pipelines are susceptible to errors in early modules
that propagate and are amplified in subsequent modules; an effect known as
cascading of errors~\citep{castro-ferreira-etal-2019-neural}.
Similarly, incorporating prior knowledge depends on the cleverness and intent
of the \emph{modeler} and generally becomes more difficult when heterogeneous
constraints are involved or the number of constraints
increases~\citep{Garbacea-etal-2022-constrained}.
Furthermore, standard text-generation architectures do not lend themselves well
for manually applying constraints. Due to the autoregressive generation of
tokens from left to right, constraints at arbitrary positions cannot be
implemented easily or only with additional
trade-offs~\citep{Garbacea-etal-2022-constrained}.
For example, end rhymes, which come at the end of a verse, cannot be
constrained in isolation due dependencies on previously generated tokens. A
commonly applied work-around for this problem is to generate each verse in
reverse~\citep{lau-etal-2018-deep,jhamtani-etal-2019-learning,van-de-cruys-2020-automatic,xue-etal-2021-deeprapper}.

In this work, we thus aim to reduce the amount of human supervision in poetry
generation and explore viable end-to-end solutions. We hypothesize that failing
to do so far has the following root causes: (i) lack of available training
data. Poetry corpora labeled with aesthetic styles are few and far between and
we speculate that they do not suffice to train a generalized model. (ii)
Unfavorable tokenization algorithms. Aesthetic styles of poetry such as rhyme,
meter, and alliteration are often expressed at the character-level while most
available off-the-shelf pre-trained models operate at the
subword-level~\citep{kudo-richardson-2018-sentencepiece}.
\citet{xue-etal-2022-byt5} showed that character-level models (also known as
\emph{token-free} models) excel at other character-level tasks so we assume
that they would perform similarly well at poetry generation. Our key
contributions are as follows:
\begin{enumerate}
  \item We pre-train \bygpt, to our knowledge the first decoder-only
    transformer for character-level language modeling.
  \item We create \quatrain, a large machine-labeled poetry corpus of quatrains
    in German and English.
  \item By fine-tuning \bygpt on \quatrain, we show that it
    learns character-level styles better than subword-based systems,
    such as \gpt and \tfive[m], as well as other token-free models like
    \tfive[By], while being more parameter efficient and also faring well
    compared to humans.
  \item We further demonstrate that \bygpt exhibits few memorization problems,
    understands poetry better than \gpt and \chatgpt, and also performs well on tasks that do not operate at the character-level.
\end{enumerate}

\section{Background}\label{sec:background} 
In formal verse poetry, poems have to follow strict patterns and rules of
language which we term \emph{styles}. Our goal is to train an end-to-end poetry
generation system which learns to adhere to specified styles by itself. We
refer to this as \emph{style-conditioned} poetry generation. In our work, we
focus on generating quatrains and conditioning on the following defining styles
of formal verse poetry (cf. Figure~\ref{fig:example-quatrain}):

\paragraph{Rhyme}
A rhyme is the repetition of the same or similar sounds in the final accented
syllables of words, which must be preceded by differing
consonants~\citep{harmon2000handbook}. If all conditions are met, we speak of
perfect rhymes, and if some of them are violated, for example, because the
final sounds are different or the words are identical, we speak of imperfect
rhymes. In a quatrain with ABAB rhyme scheme, the first and third line endings
rhyme, as do the second and fourth lines.

\paragraph{Meter}
Meter refers to the rhythmic pattern within a verse. In modern poetry, this
rhythm is usually accented-syllabic, that is, the succession of
stressed~(\metricsymbols*{_}) and unstressed syllables~(\metricsymbols*{u})
occurs at regular intervals~\citep{harmon2000handbook}. The rhythmic unit is
also known as a foot and the meter of a verse can thus be described as a
sequence of feet. In English poetry, common feet are iambic~(\metricsymbols*{u
_}), trochaic~(\metricsymbols*{_ u}), anapestic~(\metricsymbols*{u u _}), and
dactylic~(\metricsymbols*{_ u u}). For conditioning on meter, we consider all
metric feet appearing in our datasets~(cf.
Appendix~\ref{sec:corpora-statistics}).

\paragraph{Alliteration}
\citet{harmon2000handbook} define alliteration as the repetition of the same
consonant sounds or any vowel sounds at the beginning of words or syllables
that are close together in a verse. In formal verse, alliteration is
secondary to rhyme and meter, follows less strict constraints, and is therefore
not as easily classified. In this work, we thus consider the \emph{level} of
alliteration instead, which we classify as either \emph{low}, \emph{medium}, or
\emph{high} (cf. \S\ref{sec:data}).

\section{Models}\label{sec:methods} 
We induce a range of end-to-end poetry generation systems for English and
German by fine-tuning pre-trained transformer
models~\citep{vaswani2017attention}. For conditioning on style, we consider two
architectural variants---encoder-decoder
transformers~\citep{xue-etal-2021-mt5,xue-etal-2022-byt5} and decoder-only
transformers~\citep{Radford2019,brown2020fewshot}. As explained in
\S\ref{sec:introduction}, we focus on token-free models, but also consider
subword-level models for comparison. We do not experiment with models with more
than 400 million parameters since they exceed the capacity of our available GPU
resources.

\paragraph{Encoder-Decoder}
For encoder-decoder models, we initialize the encoder with a joint triple of
rhyme scheme, meter, and, alliteration level and generate a quatrain with the
decoder. We represent each style by a special token which we add to the model
vocabulary. We use \tfive[By]~\citep{xue-etal-2022-byt5}, a token-free
pre-trained encoder-decoder model, as a baseline. For comparison with
subword-level approaches, we fine-tune \tfive[m]~\citep{xue-etal-2021-mt5}.

\paragraph{Decoder-only}
As the input for encoder-decoder models is a relatively short sequence of
styles, this could lead to an underutilization of the encoder. We thus
hypothesize that a decoder-only model, with styles supplied as a prompt string,
would be better suited for our task. On the subword-level, multiple models,
such as \gpt~\citep{Radford2019}, are readily available. However, to our best
knowledge, no such model exists at the character-level yet, which is why we
train our own. Since our new model shares some similarities with the \gpt[]
family of models, but has its origin in \tfive[By] (see \S\ref{sec:bygpt}), we
refer to it as \emph{\bygpt}. An overview of all models we use can be seen in
Table~\ref{tab:models}.
\begin{table}[t]
  \centering
  \setlength{\tabcolsep}{5pt}
  \begin{tabular}[c]{llrlc}
    \toprule
    \textbf{Name} & \textbf{Size} & \textbf{Params} & \textbf{Enc/Dec} & \textbf{Token-free}\\
    \midrule
    \multirow{3}{*}{\bygpt}
    & small  & 73.5m  & Decoder & \cmark\\
    & base   & 139.2m & Decoder & \cmark\\
    & medium & 289.1m & Decoder & \cmark\\
    \midrule
    \multirow{2}{*}{\gpt}
    & base   & 124.4m & Decoder & \xmark\\
    & medium & 354.8m & Decoder & \xmark\\
    \midrule
    \tfive[By] & small  & 300m  & Enc-Dec & \cmark\\
    \midrule
    \tfive[m] & small  & 300m  & Enc-Dec & \xmark\\
    \bottomrule
  \end{tabular}
  \caption{Pre-trained models we fine-tune. \bygpt is a new model developed by
  us. The German \gpt model we use does not exist in medium
  size~\citet{Minixhofer_GerPT2_German_large_2020}, which is why we only use a
  base model there.}%
  \label{tab:models}
\end{table}

\section{\bygpt}\label{sec:bygpt}
For pre-training our own token-free decoder-only model \bygpt, we start by
modifying the architecture of \tfive[By] and discard its encoder component. We
then initialize the weights with the decoder of \tfive[By]\footnote{By
referring to this process as weight initialization (rather than continued
pre-training), we adopt the terminology of \citet{rothe-etal-2020-leveraging}.
This is intuitively sensible as we reuse only a subset of weights in a very
different architecture. Consequently, our model experiences considerably less
exposure to training data compared to its competitors during pre-training.} to
warm-start the training process~\citep{rothe-etal-2020-leveraging,tang2022mvp}.
We repeat this for the three smallest variants of \tfive[By]. Because
\tfive[By] has an asymmetrical architecture, the resulting models retain only
25\% of its parameters. We refer to their model sizes as small, base, and
medium.

\begin{figure}
  \includegraphics[width=\textwidth]{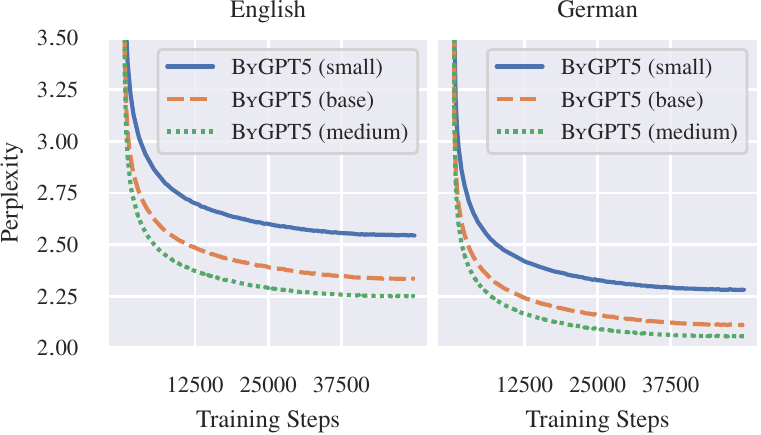}
  \caption{Perplexity on the training data when pre-training \bygpt for English
  and German.}%
  \label{fig:loss}
\end{figure}

As training data, we use \webtext~\citep{gao2021pile} for English and
\cc~\citep{conneau-etal-2020-unsupervised} for German. For hyper-parameters, we
follow \citet{Radford2019} and \citet{brown2020fewshot} and use Adam with a
weight decay of 0.1, a batch size of 512, varying learning rates depending on
model size, and train on a causal language modeling objective for 50k steps
following \citet{lester-etal-2021-power}. We provide loss curves in
Figure~\ref{fig:loss}. As might be expected, the perplexity of larger models is
generally lower than that of smaller counterparts.

\section{Datasets}\label{sec:data} 
We collect a range of labeled and unlabeled datasets of English and German
poetry~(cf. Table~\ref{tab:datasets}). As shown, we were able to procure
labeled corpora for rhyme and meter, but they are far too small to train a
poetry generation system. Instead, we use the bigger unlabeled corpora, as
training data, by labeling them automatically~\citep{belouadi2022uscore}. To
make full use of the data, we use not only real quatrains but also
\emph{pseudo-quatrains} (any consecutive sequence of four lines), amounting to
over 660k quatrains for English and 1.4m for German. We refer to this new
dataset as \quatrain, with further statistics in
Appendix~\ref{sec:corpora-statistics}. In the following, we explain the
labeling process for each style.
\begin{table}
  \begin{threeparttable}
    \begin{tabular}{llrccc}
      \toprule
      \textbf{Dataset} & \textbf{Language} & \textbf{Verses} & \textbf{R} & \textbf{M} & \textbf{A}\\
      \midrule
      \epgsf    & English &  1k & \cmark& \cmark & \xmark\\
      \prosodic & English &  2k & \xmark& \cmark & \xmark\\
      \forb     & English &  1k & \cmark& \cmark & \xmark\\
      \chicago  & English & 95k & \cmark& \xmark & \xmark\\
      \antik    & German  &  4k & \cmark& \cmark & \xmark\\
      \grc      & German  & 41k & \cmark& \xmark & \xmark\\
      \midrule
      \epg & English & 2.8m & \xmark & \xmark & \xmark\\
      \dlk & German &  2.8m & \xmark & \xmark & \xmark\\
      \midrule
      \multirow{2}{*}{\quatrain}
        & English & 2.7m\tnote{\textasteriskcentered} & \cmark\tnote{\textdagger} & \cmark\tnote{\textdagger} & \cmark\tnote{\textdagger}\\
        & German  & 5.9m\tnote{\textasteriskcentered} & \cmark\tnote{\textdagger} & \cmark\tnote{\textdagger} & \cmark\tnote{\textdagger}\\
      \bottomrule
    \end{tabular}
    \begin{tablenotes}
      \item[\textasteriskcentered]  Verses may occur in multiple pseudo-quatrains.
      \item[\textdagger]  Labels are obtained from classifiers.
    \end{tablenotes}
  \caption{Available poetry datasets for rhyme (R), meter (M), and alliteration
  (A). Unlabeled corpora (middle) are orders of magnitude larger than labeled
  corpora (top), and we label them automatically (bottom). Further information
  can be found in Appendix~\ref{sec:corpora-statistics}.}%
  \label{tab:datasets}
  \end{threeparttable}
\end{table}

For automatically labeling rhyme and meter, we leverage the available labeled
data and train a range of classifiers. We evaluate them on held-out gold data
and subsequently use the best performing classifier for each style (cf.
Appendix~\ref{sec:corpora-statistics}). Meter classification is a multiclass
classification problem with a single verse as input, while rhyme classification
is a binary classification problem with two verses separated by a special token
as input. We classify the meter of a quatrain by choosing the dominant meter
among the verses\footnote{Since in formal verse poetry, a meter is maintained
throughout a poem, this procedure is meaningful.}, and the rhyme scheme by
determining which verses rhyme and which do not.

As no readily available poetry datasets include labels for alliteration, we
approach the problem in a different way. The quantification of the level of
alliteration in a document is a long known research
problem~\citep{Skinner1939TheAI,leavitt1976allit,blain1987allit,2014bennerallit}.
Let $v_i$ be the atomic units of sound in verse $v$, \citet{blain1987allit}
quantify alliteration as
\begin{equation}
  \operatorname{allit}(v)=
  \frac{
    \sum_{i=1}^{|v|}\sum_{j=i+1}^{|v|}\frac{\operatorname{f}(v_i, v_j)}{j-i}
  }{
    \sum_{i=1}^{|v|}\sum_{j=i+1}^{|v|}\frac{1}{j-i}
  },
\end{equation}
where $\operatorname{f}(\cdot)$ is a similarity function of two sounds; the
default simply testing for equality. Intuitively, $\operatorname{allit}(\cdot)$
counts alliterative sounds in a verse, applies a distance penalty, and
normalizes the score to $[0, 1]$. To get a score for quatrains, we average the
alliteration level of all verses. We consider initial phonemes of words, as
well as all further stressed phonemes as atomic sound units $v_i$, and to
determine phonemes and stress, we use a grapheme-to-phoneme conversion
model~\citep{zhu22_interspeech}. Further, we conduct an internal study to
determine several intensity thresholds based on a sample of quatrains. We
classify the alliteration level of a quatrain as \emph{low} if the score is
below 0.05, \emph{medium} if the score is below 0.1, and \emph{high} if it is
above that.

\section{Experiments}\label{sec:experiments} 
For fine-tuning, we use the same hyperparameters as in \S\ref{sec:bygpt} for
all models, but reduce the batch size to 128 (for efficiency reasons). We
induce separate models for each language in \quatrain and train for 10
epochs.\footnotemark{} We
conduct both automatic (\S\ref{sec:auto-eval}) and human evaluation
(\S\ref{sec:human-eval}). Examples of generated quatrains can be found in
Appendix~\ref{sec:example-quatrains}.

\subsection{Automatic Evaluation}\label{sec:auto-eval}
\begin{figure*}
  \centering
  \includegraphics[width=\textwidth]{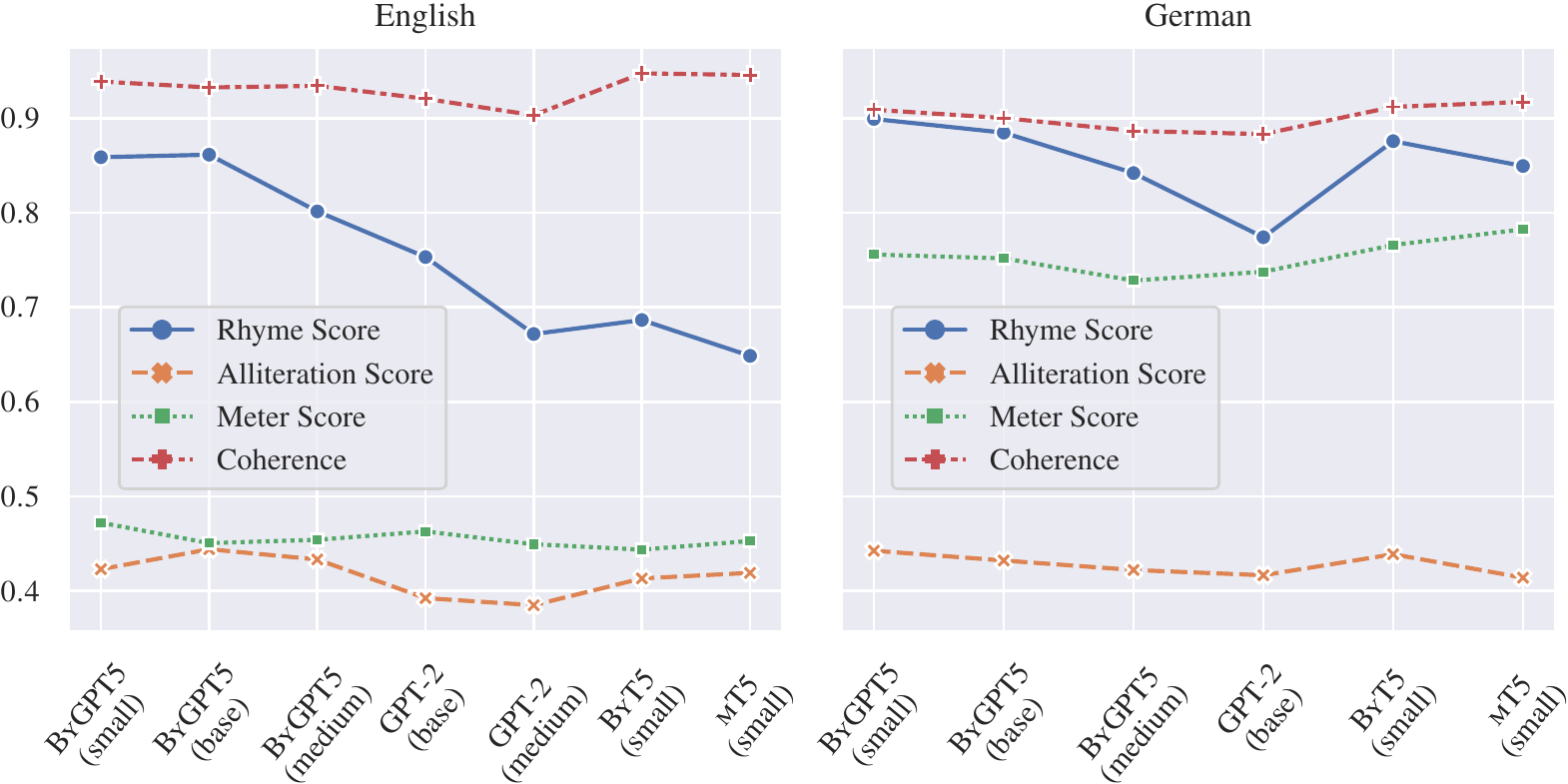}
  \caption{Automatic evaluation results for all models on English and German.}%
  \label{fig:scores}
\end{figure*}
For automatic evaluation, we select four common rhyme schemes (AABB, ABAB,
ABBA, and ABCB), the most popular meters per language (iambus, trochee,
anapest, and dactyl for English; iambus, trochee, and alexandrine for German),
and all levels of alliteration to create 75 poems per model for each possible
combination. To find out if styles are properly reflected in generated
quatrains we reuse the classifiers from \S\ref{sec:data}, i.e., we use them to
classify the generated poems and see if the styles match. We define the
following metrics:
\begin{description}
  \item[Rhyme Score] computes the recall of verses that should rhyme in a
    quatrain, as well as the recall of verses that should not and takes their
    arithmetic average.
  \item[Alliteration Score] is 1 if a quatrain has the correct alliteration
    level, else 0.
  \item[Meter Score] is the fraction of verses with correctly classified meters.
  \item[Coherence] uses \bert for next sentence
    prediction~\citep{devlin-etal-2019-bert} to assess discourse relations of
    verses~\citep{duari-2021-ffcd,shi-demberg-2019-next}. The score is the
    fraction of consecutive verse pairs that are correctly classified to come
    after one another.
\end{description}
\footnotetext{ All models converge within the designated number of epochs, and
throughout each epoch, the ordering of the systems remains consistent with our
final results from automatic evaluation.}

We provide the scores for each model averaged over all generated quatrains
(2700 for German and 3600 for English) in Figure~\ref{fig:scores}. Although all
models manage to learn to follow aesthetic styles to some degree, there are
noticeable score differences.

On rhyme, all \bygpt variants collectively outperform all \gpt models on both
languages by 5\%-20\%. Similarly, \tfive[By] consistently outperforms \tfive[m]
by \textasciitilde5\%. This supports our theory that token-free models are
better suited for character-level styles. Further, \bygpt (small) performs
2\%-15\%  better than \tfive[By] (small) which means we can discard the encoder
while still improving performance. Surprisingly though, base \bygpt and \gpt
achieve higher scores than their medium variants. While this may initially
suggest that larger decoders prioritize (meaningful) content, whereas smaller
decoders focus on style, the high coherence seen across all models weakens this
hypothesis. Instead, we speculate that this may be an overfitting problem. In
particular, smaller models, up to base size, may be better suited for
generating shorter texts such as quatrains. Another surprising finding is that
\tfive[By] (small) performs worse than \gpt (base) on English. We investigate
this further in \S\ref{sec:human-eval}.

In terms of meter, all models perform very similar to one another. Whereas
\bygpt (small) performs best on English by a small margin, it is outperformed
by \tfive[m] (small) on German. This result is not surprising. Since meter is a
syllable-level style, subword-level language models also manage to pick it up
reasonably well~\citep{lau-etal-2018-deep}. Interestingly though, on English
the scores are much lower overall than on German. A reason for this may be that
the occurrence of different meters is much more evenly distributed in German
\quatrain (cf.\ Table~\ref{tab:quatrain-stats} in the appendix). While in
German only about 60\% of all meters are iambs, in English it is over 80\%,
making it difficult for models to learn other meters. We identify further
reasons in \S\ref{sec:human-eval}.

Alliteration is the style that all models are the worst at. Our formulation of
alliteration levels may make it difficult for models to pick up its semantics.
Still, \bygpt (base) performs the best on English, and \bygpt (small) and
\tfive[By] (small) perform the best on German, suggesting that token-free
models have an advantage on this style.

In general, small and base \bygpt perform the best on all three styles in
English and on two of them in German. It appears that they have an advantage in
terms of tokenization algorithms (outperforming subword-level \gpt and
\tfive[m]), architecture (outperforming encoder-decoder \tfive[By]), and size
(the medium variant performs worse).

\subsection{Human Evaluation}\label{sec:human-eval}
\begin{figure*}
  \centering
  \includegraphics[width=\textwidth]{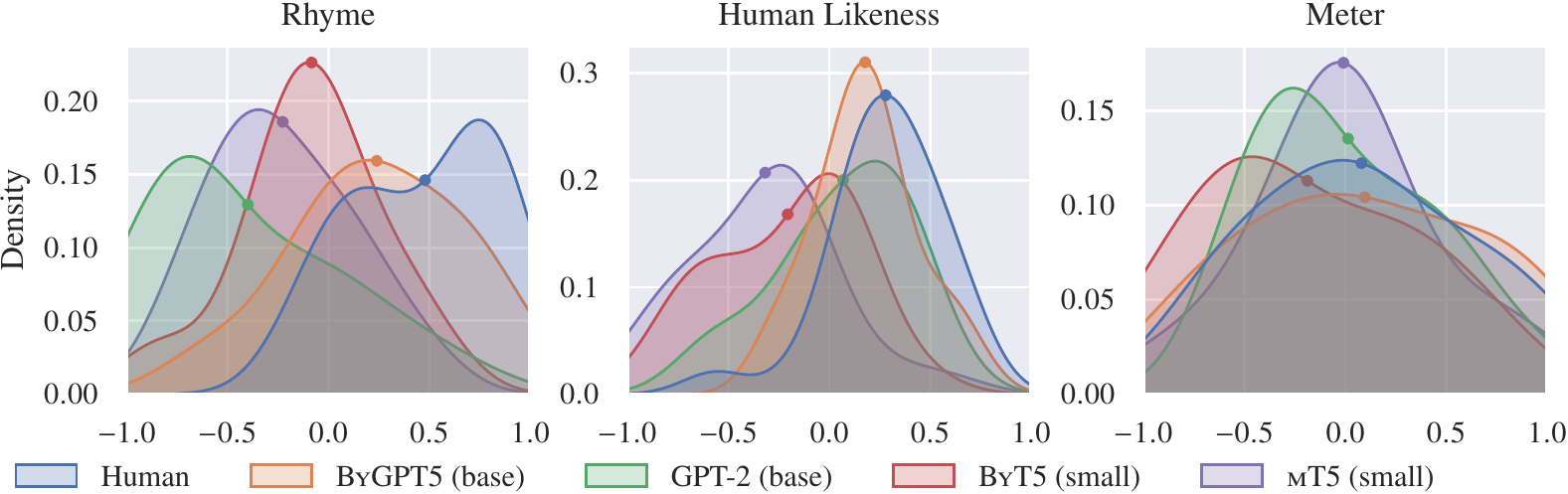}
  \caption{Distributions of BWS scores for rhyme, human likeness, and meter
  annotations through kernel density estimation. Scores range from -1 (very
  bad) to 1 (very good). The \enquote{\textbullet} markers denote expected
  values.}%
  \label{fig:kde}
\end{figure*}
To further validate the effectiveness of our models, we conduct a human
evaluation campaign using \emph{best-worst scaling} (BWS) as a means of
annotation~\citep{louviere_flynn_marley_2015}. BWS is a variant of comparative
annotation that produces high-quality results while keeping the number of
required annotations
low~\citep{kiritchenko-mohammad-2016-capturing,kiritchenko-mohammad-2017-best}.
Annotators are presented with tuples of $n$ items (usually $n=4$) and asked to
identify the best and worst item based on a specified property. By subtracting
the fraction of times an item is chosen as the best from the fraction of times
it is chosen as the worst, real-valued scores ranging from -1 (bad) to 1 (good)
can be obtained~\citep{Orme2009MaxDiffA}.

In our annotations, we consider three properties: rhyme, meter, and human
likeness, i.e., the likelihood of a poem being written by a human. We exclude
alliteration to reduce the workload on our annotators. Similarly, we
exclusively evaluate on English (cf. Appendix~\ref{sec:demographics} for a
small-scale evaluation in German) and only consider the top-performing
model within each model class based on the results of automatic evaluation. The
models in question thus are \bygpt (base), \gpt (base), \tfive[By] (small), and
\tfive[m] (small). Furthermore, we only choose from three rhyme schemes (AABB,
ABAB and ABBA), two meters (iambus and trochee), and one level of alliteration
(medium), and create four poems per system for each possible combination. In
addition, we also randomly sample human quatrains from our datasets that match
the constraints and create 120 4-tuples from the combined set of quatrains.

Four annotators then annotate rhyme and human likeness, whereas meter is
evaluated by a single expert annotator only (cf.
Appendix~\ref{sec:demographics}). Since we have multiple annotators working on
rhyme and human likeness we use the \emph{split-half reliability} (SHR)
measure~\citep{kiritchenko-mohammad-2017-best} to assess their consistency. SHR
is calculated by splitting the annotations into two sets, computing scores for
each set, and then computing their Spearman rank correlation coefficient.

Figure~\ref{fig:kde} displays a kernel density estimate for each property, with
distributions shifted to the right indicating better performance. On rhymes, we
obtain an SHR of $\rho = 0.77$ which demonstrates a high agreement between
annotators. Human rhymes are ranked the highest overall, whereas \bygpt comes
in as a close second, followed by \tfive[By]. \tfive[m] and \gpt perform the
worst. This is a bit different from our findings during automatic evaluation
where \gpt (base) was ranked higher than \tfive[By] (small) on English. An
analysis of \gpt generated quatrains revealed a predominance of imperfect
rhymes as a likely cause. As our rhyme classifier is trained on binary labels
it is unable to detect this, but human annotators perceive this kind of rhyme
as worse.

With $\rho = 0.54$, the SRH of human likeness is noticeably lower than for
rhyme. On the one hand, this suggests that this task could be more subjective;
on the other hand, the generated quatrains must be sufficiently human-like for
subjectivity to be a factor. Indeed, although humans rank higher than \bygpt
which in turn ranks higher than \gpt, they all perform noticeably more similar
than for rhyme. Nonetheless, we can observe that \tfive[By], and especially
\tfive[m] rank a bit lower. Both models were pre-trained on corrupted spans and
have thus never seen truly natural text during
pre-training~\citep{zhu22_interspeech,2020t5,lester-etal-2021-power} which we
believe could be a possible cause.

The distributions for meter have large variances for all models, and also
humans. This is surprising, as it implies that our annotator does not think
that humans are superior, even though the automatic evaluation of English
models was not particularly strong on meter. We hypothesize that even among
real English poets, there is a significant amount of poetry that does not
strictly adhere to metric constraints, so language models only learn to follow
them freely as well. Nonetheless, we can still see that, among models, \bygpt
is rated highest, followed by \gpt, \tfive[m], and lastly \tfive[By],
reflecting our findings of automatic evaluation. Interestingly, \bygpt also
ranks higher than humans.

Overall, our human evaluation suggests that \bygpt performs best across all
properties evaluated, which is consistent with our automatic evaluation.
Moreover, \bygpt has shown the ability to perform at a level comparable to
humans, and even surpass human performance in the meter property.

\section{Analysis}\label{sec:analysis} 
We continue with a deeper analysis and look into low-resource training
(\S\ref{sec:low-res}), quantify memorization (\S\ref{sec:memo}), evaluate the
performance of token-free models on non-character-based, high-level tasks
(\S\ref{sec:doc-lvl}), introspect the models' understanding of style when
predicting tokens (\S\ref{sec:tok-attr}), and compare \bygpt with \chatgpt
(\S\ref{sec:chatgpt}).

\subsection{Low-resource Training}\label{sec:low-res}
\begin{figure}
  \centering
  \includegraphics{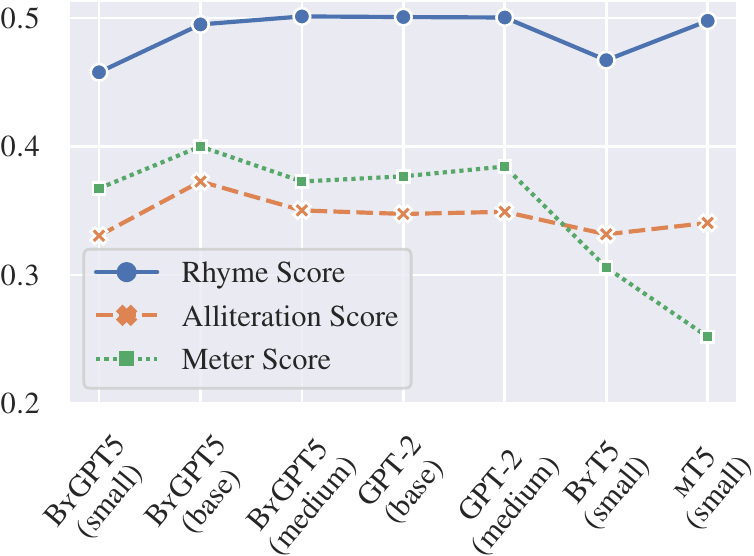}
  \caption{Automatic evaluation of low-resource models.}%
  \label{fig:low-res}
\end{figure}

We hypothesized that a large training corpus is an important factor in
successfully training an end-to-end poetry generation system. We examine this
hypothesis by selecting a 5\% subset of English \quatrain (33k quatrains) and
re-training our models using the same hyperparameters as in
\S\ref{sec:experiments}. Figure~\ref{fig:low-res} shows how well these new
low-resource models adhere to style constraints, similar to the automatic
evaluation of full training in Figure~\ref{fig:scores}.

Compared to full training, all low-resource models are noticeably worse at
adhering to style. Specifically, the performance drops by 15\%-40\% for rhyme,
5\%-20\% for meter, and 5\%-10\% for alliteration. In addition, the overall
performance difference between all models is much smaller than in
\S\ref{sec:auto-eval}. While these findings support our hypothesis that
training on large datasets is essential, they also reveal that \bygpt
demonstrates the largest improvements as the dataset size increases (cf.
Figure~\ref{fig:scores}). We therefore theorize that larger datasets lead to
substantial performance gains in poetry generation \emph{only} when coupled
with architectures that excel at character-level tasks.

Nevertheless, even in low-resource scenarios, \bygpt (base) outperforms the
other models in all categories except rhyme, where a few other systems perform
similarly. This suggests that the conclusions drawn in \S\ref{sec:auto-eval}
hold to some extent even when the available training data is limited.

\subsection{Extractive Memorization}\label{sec:memo}
A common problem of language models, known as extractive memorization (EM), is
generating verbatim copies from the training data during
inference~\citep{carlini2022memo,raunak-etal-finding-memo,pmlr-v108-meehan20a}.
According to \citet{carlini2022memo} EM occurs when a language model's
continuation of a string is part of the data it was trained
on.
Since the inputs to our language models are strings of style, this formulation
lends itself well to our case: to detect memorization we simply have to check
if generated poems appear in \quatrain. In
Table~\ref{tab:memorization-emotion}, we compute the EM rates of the quatrains
generated in \S\ref{sec:auto-eval}. To account for negligible variations, we do
not compare raw strings, but calculate the Ratcliff-Obershelp
similarity~\citep{ratcliff1988gestalt}, and assume that two strings are
equivalent if their similarity exceeds 0.7.
\begin{table}
  \centering
  \begin{tabular}[c]{lrrr}
    \toprule

    & \multicolumn{2}{c}{\bf Memorization} & \textbf{Emotion}\\
    \cmidrule(r){2-3}\cmidrule(l){4-4}
    \textbf{Model} & English & German & German \\
    \midrule
    \bygpt (small)  &  0.0\% &  0.0\% & 0.676\\
    \bygpt (base)   &  0.0\% & 0.04\% & 0.680\\
    \bygpt (medium) &  0.0\% & 0.81\% & 0.659\\
    \midrule
    \gpt (base)     &      0.39\% & \hl{1.81\%} & 0.676\\
    \gpt (medium)   & \hl{3.64\%} &         --- &   ---\\
    \midrule
    \tfive[By]      & 0.0\% & 0.0\%& 0.691\\
    \midrule
    \tfive[m]       & 0.0\% & 0.0\%& \hl{0.696}\\
    \bottomrule
  \end{tabular}
  \caption{Extractive memorization rates (English \& German) and recall
  on emotion generation (German).}%
  \label{tab:memorization-emotion}
\end{table}

As can be seen, \gpt suffers from memorization the most. On English, over 3\%
of all outputs of \gpt (medium) are copied. While \bygpt also copies quatrains
to an extent, it is much less affected in comparison. On English, \bygpt
(medium) does not copy anything and on German only 0.81\% of all outputs. As a
general trend, we can see that bigger models tend to copy more data than
smaller ones---a finding shared by others~\citep{carlini2022memo}.
Interestingly, the encoder-decoder models \tfive[By] and \tfive[m] do not seem
to be affected by this problem at all, most likely because styles are not used
as a prompt, but are fed into the encoder separately.

\subsection{Higher-level Styles}\label{sec:doc-lvl}
We also explore how token-free models perform on higher-level styles which are
not character- or subword-level phenomena. In particular, we focus on emotion
using \poemo~\citep{haider-etal-2020-po}, a dataset of eight aesthetic emotions
in poetry (cf. Appendix~\ref{sec:corpora-statistics}). By conditioning our
models on these emotions, we can assess their ability to understand and depict
emotion in poetry.

As in \S\ref{sec:data}, we leverage automatic labeling. To that end, we train a
classifier on German \poemo as in \citet{haider-etal-2020-po} and reproduce the
results. We then classify emotions in German \quatrain and retrain our systems
by conditioning them on all emotions in a quatrain. To evaluate how well the
models can discriminate emotions, we condition them on every possible tuple of
two distinct emotions and generate 100 poems each (2800 in total), and report
the recall of correctly classified emotions.
The results in Table~\ref{tab:memorization-emotion} show that encoder-decoder
models score highest, with \tfive[m] performing best. During training,
conditioning inputs can be long and variable in size, a scenario for which
encoder-decoders may be better suited. Still, decoder-only models are not far
behind. Especially \bygpt fares well against \gpt, suggesting that token-free
models are also competitive on higher-level tasks.

\subsection{Token-level Attributions}\label{sec:tok-attr}
\begin{figure}
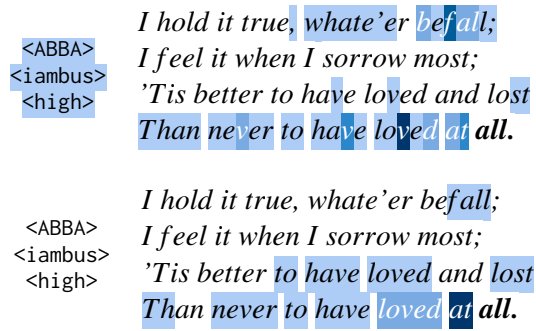


\ifx\versesize\undefined\def\versesize{\normalsize}\fi%
\ifx\stylesize\undefined\def\stylesize{\small}\fi%
{\versesize\fboxsep0pt{}
  \begin{tabular}{cl}
    \multirow{4}{*}{\stylesize\makecell{\colorbox[HTML]{ADCCF6}{\textcolor{black}{\strut{}{\texttt{<ABBA>}}}}\\\colorbox[HTML]{ADCCF6}{\textcolor{black}{\strut{}{\texttt{<iambus>}}}}\\\colorbox[HTML]{ADCCF6}{\textcolor{black}{\strut{}{\texttt{<high>}}}}}}
    & \colorbox[HTML]{FFFFFF}{\textcolor{black}{\strut{}\textrm{\textit{I}}}} \colorbox[HTML]{FFFFFF}{\textcolor{black}{\strut{}\textrm{\textit{h}}}}\colorbox[HTML]{FFFFFF}{\textcolor{black}{\strut{}\textrm{\textit{o}}}}\colorbox[HTML]{FFFFFF}{\textcolor{black}{\strut{}\textrm{\textit{l}}}}\colorbox[HTML]{FFFFFF}{\textcolor{black}{\strut{}\textrm{\textit{d}}}} \colorbox[HTML]{FFFFFF}{\textcolor{black}{\strut{}\textrm{\textit{i}}}}\colorbox[HTML]{FFFFFF}{\textcolor{black}{\strut{}\textrm{\textit{t}}}} \colorbox[HTML]{FFFFFF}{\textcolor{black}{\strut{}\textrm{\textit{t}}}}\colorbox[HTML]{FFFFFF}{\textcolor{black}{\strut{}\textrm{\textit{r}}}}\colorbox[HTML]{FFFFFF}{\textcolor{black}{\strut{}\textrm{\textit{u}}}}\colorbox[HTML]{FFFFFF}{\textcolor{black}{\strut{}\textrm{\textit{e}}}}\colorbox[HTML]{ADCCF6}{\textcolor{black}{\strut{}\textrm{\textit{,}}}} \colorbox[HTML]{ADCCF6}{\textcolor{black}{\strut{}\textrm{\textit{w}}}}\colorbox[HTML]{ADCCF6}{\textcolor{black}{\strut{}\textrm{\textit{h}}}}\colorbox[HTML]{ADCCF6}{\textcolor{black}{\strut{}\textrm{\textit{a}}}}\colorbox[HTML]{ADCCF6}{\textcolor{black}{\strut{}\textrm{\textit{t}}}}\colorbox[HTML]{ADCCF6}{\textcolor{black}{\strut{}\textrm{\textit{e}}}}\colorbox[HTML]{ADCCF6}{\textcolor{black}{\strut{}\textrm{\textit{'}}}}\colorbox[HTML]{ADCCF6}{\textcolor{black}{\strut{}\textrm{\textit{e}}}}\colorbox[HTML]{FFFFFF}{\textcolor{black}{\strut{}\textrm{\textit{r}}}} \colorbox[HTML]{79ABE2}{\textcolor{white}{\strut{}\textrm{\textit{b}}}}\colorbox[HTML]{ADCCF6}{\textcolor{black}{\strut{}\textrm{\textit{e}}}}\colorbox[HTML]{005D9A}{\textcolor{white}{\strut{}\textrm{\textit{f}}}}\colorbox[HTML]{79ABE2}{\textcolor{white}{\strut{}\textrm{\textit{a}}}}\colorbox[HTML]{79ABE2}{\textcolor{white}{\strut{}\textrm{\textit{l}}}}\colorbox[HTML]{ADCCF6}{\textcolor{black}{\strut{}\textrm{\textit{l}}}}\colorbox[HTML]{ADCCF6}{\textcolor{black}{\strut{}\textrm{\textit{;}}}}\\
    & \colorbox[HTML]{FFFFFF}{\textcolor{black}{\strut{}\textrm{\textit{I}}}} \colorbox[HTML]{FFFFFF}{\textcolor{black}{\strut{}\textrm{\textit{f}}}}\colorbox[HTML]{FFFFFF}{\textcolor{black}{\strut{}\textrm{\textit{e}}}}\colorbox[HTML]{FFFFFF}{\textcolor{black}{\strut{}\textrm{\textit{e}}}}\colorbox[HTML]{FFFFFF}{\textcolor{black}{\strut{}\textrm{\textit{l}}}} \colorbox[HTML]{FFFFFF}{\textcolor{black}{\strut{}\textrm{\textit{i}}}}\colorbox[HTML]{FFFFFF}{\textcolor{black}{\strut{}\textrm{\textit{t}}}} \colorbox[HTML]{FFFFFF}{\textcolor{black}{\strut{}\textrm{\textit{w}}}}\colorbox[HTML]{FFFFFF}{\textcolor{black}{\strut{}\textrm{\textit{h}}}}\colorbox[HTML]{FFFFFF}{\textcolor{black}{\strut{}\textrm{\textit{e}}}}\colorbox[HTML]{FFFFFF}{\textcolor{black}{\strut{}\textrm{\textit{n}}}} \colorbox[HTML]{FFFFFF}{\textcolor{black}{\strut{}\textrm{\textit{I}}}} \colorbox[HTML]{FFFFFF}{\textcolor{black}{\strut{}\textrm{\textit{s}}}}\colorbox[HTML]{FFFFFF}{\textcolor{black}{\strut{}\textrm{\textit{o}}}}\colorbox[HTML]{FFFFFF}{\textcolor{black}{\strut{}\textrm{\textit{r}}}}\colorbox[HTML]{FFFFFF}{\textcolor{black}{\strut{}\textrm{\textit{r}}}}\colorbox[HTML]{FFFFFF}{\textcolor{black}{\strut{}\textrm{\textit{o}}}}\colorbox[HTML]{FFFFFF}{\textcolor{black}{\strut{}\textrm{\textit{w}}}} \colorbox[HTML]{FFFFFF}{\textcolor{black}{\strut{}\textrm{\textit{m}}}}\colorbox[HTML]{FFFFFF}{\textcolor{black}{\strut{}\textrm{\textit{o}}}}\colorbox[HTML]{FFFFFF}{\textcolor{black}{\strut{}\textrm{\textit{s}}}}\colorbox[HTML]{FFFFFF}{\textcolor{black}{\strut{}\textrm{\textit{t}}}}\colorbox[HTML]{FFFFFF}{\textcolor{black}{\strut{}\textrm{\textit{;}}}}\\
    & \colorbox[HTML]{FFFFFF}{\textcolor{black}{\strut{}\textrm{\textit{'}}}}\colorbox[HTML]{FFFFFF}{\textcolor{black}{\strut{}\textrm{\textit{T}}}}\colorbox[HTML]{FFFFFF}{\textcolor{black}{\strut{}\textrm{\textit{i}}}}\colorbox[HTML]{FFFFFF}{\textcolor{black}{\strut{}\textrm{\textit{s}}}} \colorbox[HTML]{FFFFFF}{\textcolor{black}{\strut{}\textrm{\textit{b}}}}\colorbox[HTML]{FFFFFF}{\textcolor{black}{\strut{}\textrm{\textit{e}}}}\colorbox[HTML]{FFFFFF}{\textcolor{black}{\strut{}\textrm{\textit{t}}}}\colorbox[HTML]{FFFFFF}{\textcolor{black}{\strut{}\textrm{\textit{t}}}}\colorbox[HTML]{FFFFFF}{\textcolor{black}{\strut{}\textrm{\textit{e}}}}\colorbox[HTML]{FFFFFF}{\textcolor{black}{\strut{}\textrm{\textit{r}}}} \colorbox[HTML]{FFFFFF}{\textcolor{black}{\strut{}\textrm{\textit{t}}}}\colorbox[HTML]{FFFFFF}{\textcolor{black}{\strut{}\textrm{\textit{o}}}} \colorbox[HTML]{FFFFFF}{\textcolor{black}{\strut{}\textrm{\textit{h}}}}\colorbox[HTML]{FFFFFF}{\textcolor{black}{\strut{}\textrm{\textit{a}}}}\colorbox[HTML]{ADCCF6}{\textcolor{black}{\strut{}\textrm{\textit{v}}}}\colorbox[HTML]{FFFFFF}{\textcolor{black}{\strut{}\textrm{\textit{e}}}} \colorbox[HTML]{FFFFFF}{\textcolor{black}{\strut{}\textrm{\textit{l}}}}\colorbox[HTML]{FFFFFF}{\textcolor{black}{\strut{}\textrm{\textit{o}}}}\colorbox[HTML]{ADCCF6}{\textcolor{black}{\strut{}\textrm{\textit{v}}}}\colorbox[HTML]{FFFFFF}{\textcolor{black}{\strut{}\textrm{\textit{e}}}}\colorbox[HTML]{FFFFFF}{\textcolor{black}{\strut{}\textrm{\textit{d}}}} \colorbox[HTML]{FFFFFF}{\textcolor{black}{\strut{}\textrm{\textit{a}}}}\colorbox[HTML]{FFFFFF}{\textcolor{black}{\strut{}\textrm{\textit{n}}}}\colorbox[HTML]{FFFFFF}{\textcolor{black}{\strut{}\textrm{\textit{d}}}} \colorbox[HTML]{FFFFFF}{\textcolor{black}{\strut{}\textrm{\textit{l}}}}\colorbox[HTML]{FFFFFF}{\textcolor{black}{\strut{}\textrm{\textit{o}}}}\colorbox[HTML]{ADCCF6}{\textcolor{black}{\strut{}\textrm{\textit{s}}}}\colorbox[HTML]{ADCCF6}{\textcolor{black}{\strut{}\textrm{\textit{t}}}}\\
    & \colorbox[HTML]{ADCCF6}{\textcolor{black}{\strut{}\textrm{\textit{T}}}}\colorbox[HTML]{ADCCF6}{\textcolor{black}{\strut{}\textrm{\textit{h}}}}\colorbox[HTML]{ADCCF6}{\textcolor{black}{\strut{}\textrm{\textit{a}}}}\colorbox[HTML]{ADCCF6}{\textcolor{black}{\strut{}\textrm{\textit{n}}}} \colorbox[HTML]{ADCCF6}{\textcolor{black}{\strut{}\textrm{\textit{n}}}}\colorbox[HTML]{ADCCF6}{\textcolor{black}{\strut{}\textrm{\textit{e}}}}\colorbox[HTML]{79ABE2}{\textcolor{white}{\strut{}\textrm{\textit{v}}}}\colorbox[HTML]{ADCCF6}{\textcolor{black}{\strut{}\textrm{\textit{e}}}}\colorbox[HTML]{ADCCF6}{\textcolor{black}{\strut{}\textrm{\textit{r}}}} \colorbox[HTML]{ADCCF6}{\textcolor{black}{\strut{}\textrm{\textit{t}}}}\colorbox[HTML]{ADCCF6}{\textcolor{black}{\strut{}\textrm{\textit{o}}}} \colorbox[HTML]{ADCCF6}{\textcolor{black}{\strut{}\textrm{\textit{h}}}}\colorbox[HTML]{ADCCF6}{\textcolor{black}{\strut{}\textrm{\textit{a}}}}\colorbox[HTML]{2C86CA}{\textcolor{white}{\strut{}\textrm{\textit{v}}}}\colorbox[HTML]{ADCCF6}{\textcolor{black}{\strut{}\textrm{\textit{e}}}} \colorbox[HTML]{ADCCF6}{\textcolor{black}{\strut{}\textrm{\textit{l}}}}\colorbox[HTML]{ADCCF6}{\textcolor{black}{\strut{}\textrm{\textit{o}}}}\colorbox[HTML]{00366C}{\textcolor{white}{\strut{}\textrm{\textit{v}}}}\colorbox[HTML]{ADCCF6}{\textcolor{black}{\strut{}\textrm{\textit{e}}}}\colorbox[HTML]{79ABE2}{\textcolor{white}{\strut{}\textrm{\textit{d}}}} \colorbox[HTML]{79ABE2}{\textcolor{white}{\strut{}\textrm{\textit{a}}}}\colorbox[HTML]{2C86CA}{\textcolor{white}{\strut{}\textrm{\textit{t}}}} \colorbox[HTML]{FFFFFF}{\textcolor{black}{\strut{}\textrm{\textit{\textbf{a}}}}}\colorbox[HTML]{FFFFFF}{\textcolor{black}{\strut{}\textrm{\textit{\textbf{l}}}}}\colorbox[HTML]{FFFFFF}{\textcolor{black}{\strut{}\textrm{\textit{\textbf{l}}}}}\colorbox[HTML]{FFFFFF}{\textcolor{black}{\strut{}\textrm{\textit{\textbf{.}}}}}
  \end{tabular}
}%
\global\let\stylesize\undefined%
\global\let\versesize\undefined%

  \bigskip

\ifx\versesize\undefined\def\versesize{\normalsize}\fi%
\ifx\stylesize\undefined\def\stylesize{\small}\fi%
{\versesize\fboxsep0pt{}
  \begin{tabular}{cl}
    \multirow{4}{*}{\stylesize\makecell{\colorbox[HTML]{FFFFFF}{\textcolor{black}{\strut{}{\texttt{<ABBA>}}}}\\\colorbox[HTML]{FFFFFF}{\textcolor{black}{\strut{}{\texttt{<iambus>}}}}\\\colorbox[HTML]{FFFFFF}{\textcolor{black}{\strut{}{\texttt{<high>}}}}}}
    & \colorbox[HTML]{FFFFFF}{\textcolor{black}{\strut{}\textrm{\textit{I}}}} \colorbox[HTML]{FFFFFF}{\textcolor{black}{\strut{}\textrm{\textit{h}}}}\colorbox[HTML]{FFFFFF}{\textcolor{black}{\strut{}\textrm{\textit{o}}}}\colorbox[HTML]{FFFFFF}{\textcolor{black}{\strut{}\textrm{\textit{l}}}}\colorbox[HTML]{FFFFFF}{\textcolor{black}{\strut{}\textrm{\textit{d}}}} \colorbox[HTML]{FFFFFF}{\textcolor{black}{\strut{}\textrm{\textit{i}}}}\colorbox[HTML]{FFFFFF}{\textcolor{black}{\strut{}\textrm{\textit{t}}}} \colorbox[HTML]{FFFFFF}{\textcolor{black}{\strut{}\textrm{\textit{t}}}}\colorbox[HTML]{FFFFFF}{\textcolor{black}{\strut{}\textrm{\textit{r}}}}\colorbox[HTML]{FFFFFF}{\textcolor{black}{\strut{}\textrm{\textit{u}}}}\colorbox[HTML]{FFFFFF}{\textcolor{black}{\strut{}\textrm{\textit{e}}}}\colorbox[HTML]{FFFFFF}{\textcolor{black}{\strut{}\textrm{\textit{,}}}} \colorbox[HTML]{FFFFFF}{\textcolor{black}{\strut{}\textrm{\textit{w}}}}\colorbox[HTML]{FFFFFF}{\textcolor{black}{\strut{}\textrm{\textit{h}}}}\colorbox[HTML]{FFFFFF}{\textcolor{black}{\strut{}\textrm{\textit{a}}}}\colorbox[HTML]{FFFFFF}{\textcolor{black}{\strut{}\textrm{\textit{t}}}}\colorbox[HTML]{FFFFFF}{\textcolor{black}{\strut{}\textrm{\textit{e}}}}\colorbox[HTML]{FFFFFF}{\textcolor{black}{\strut{}\textrm{\textit{'}}}}\colorbox[HTML]{FFFFFF}{\textcolor{black}{\strut{}\textrm{\textit{e}}}}\colorbox[HTML]{FFFFFF}{\textcolor{black}{\strut{}\textrm{\textit{r}}}} \colorbox[HTML]{FFFFFF}{\textcolor{black}{\strut{}\textrm{\textit{b}}}}\colorbox[HTML]{FFFFFF}{\textcolor{black}{\strut{}\textrm{\textit{e}}}}\colorbox[HTML]{ADCCF6}{\textcolor{black}{\strut{}\textrm{\textit{f}}}}\colorbox[HTML]{ADCCF6}{\textcolor{black}{\strut{}\textrm{\textit{a}}}}\colorbox[HTML]{ADCCF6}{\textcolor{black}{\strut{}\textrm{\textit{l}}}}\colorbox[HTML]{ADCCF6}{\textcolor{black}{\strut{}\textrm{\textit{l}}}}\colorbox[HTML]{FFFFFF}{\textcolor{black}{\strut{}\textrm{\textit{;}}}}\\
    & \colorbox[HTML]{FFFFFF}{\textcolor{black}{\strut{}\textrm{\textit{I}}}} \colorbox[HTML]{FFFFFF}{\textcolor{black}{\strut{}\textrm{\textit{f}}}}\colorbox[HTML]{FFFFFF}{\textcolor{black}{\strut{}\textrm{\textit{e}}}}\colorbox[HTML]{FFFFFF}{\textcolor{black}{\strut{}\textrm{\textit{e}}}}\colorbox[HTML]{FFFFFF}{\textcolor{black}{\strut{}\textrm{\textit{l}}}} \colorbox[HTML]{FFFFFF}{\textcolor{black}{\strut{}\textrm{\textit{i}}}}\colorbox[HTML]{FFFFFF}{\textcolor{black}{\strut{}\textrm{\textit{t}}}} \colorbox[HTML]{FFFFFF}{\textcolor{black}{\strut{}\textrm{\textit{w}}}}\colorbox[HTML]{FFFFFF}{\textcolor{black}{\strut{}\textrm{\textit{h}}}}\colorbox[HTML]{FFFFFF}{\textcolor{black}{\strut{}\textrm{\textit{e}}}}\colorbox[HTML]{FFFFFF}{\textcolor{black}{\strut{}\textrm{\textit{n}}}} \colorbox[HTML]{FFFFFF}{\textcolor{black}{\strut{}\textrm{\textit{I}}}} \colorbox[HTML]{FFFFFF}{\textcolor{black}{\strut{}\textrm{\textit{s}}}}\colorbox[HTML]{FFFFFF}{\textcolor{black}{\strut{}\textrm{\textit{o}}}}\colorbox[HTML]{FFFFFF}{\textcolor{black}{\strut{}\textrm{\textit{r}}}}\colorbox[HTML]{FFFFFF}{\textcolor{black}{\strut{}\textrm{\textit{r}}}}\colorbox[HTML]{FFFFFF}{\textcolor{black}{\strut{}\textrm{\textit{o}}}}\colorbox[HTML]{FFFFFF}{\textcolor{black}{\strut{}\textrm{\textit{w}}}} \colorbox[HTML]{FFFFFF}{\textcolor{black}{\strut{}\textrm{\textit{m}}}}\colorbox[HTML]{FFFFFF}{\textcolor{black}{\strut{}\textrm{\textit{o}}}}\colorbox[HTML]{FFFFFF}{\textcolor{black}{\strut{}\textrm{\textit{s}}}}\colorbox[HTML]{FFFFFF}{\textcolor{black}{\strut{}\textrm{\textit{t}}}}\colorbox[HTML]{FFFFFF}{\textcolor{black}{\strut{}\textrm{\textit{;}}}}\\
    & \colorbox[HTML]{FFFFFF}{\textcolor{black}{\strut{}\textrm{\textit{'}}}}\colorbox[HTML]{FFFFFF}{\textcolor{black}{\strut{}\textrm{\textit{T}}}}\colorbox[HTML]{FFFFFF}{\textcolor{black}{\strut{}\textrm{\textit{i}}}}\colorbox[HTML]{FFFFFF}{\textcolor{black}{\strut{}\textrm{\textit{s}}}} \colorbox[HTML]{FFFFFF}{\textcolor{black}{\strut{}\textrm{\textit{b}}}}\colorbox[HTML]{FFFFFF}{\textcolor{black}{\strut{}\textrm{\textit{e}}}}\colorbox[HTML]{FFFFFF}{\textcolor{black}{\strut{}\textrm{\textit{t}}}}\colorbox[HTML]{FFFFFF}{\textcolor{black}{\strut{}\textrm{\textit{t}}}}\colorbox[HTML]{FFFFFF}{\textcolor{black}{\strut{}\textrm{\textit{e}}}}\colorbox[HTML]{FFFFFF}{\textcolor{black}{\strut{}\textrm{\textit{r}}}} \colorbox[HTML]{ADCCF6}{\textcolor{black}{\strut{}\textrm{\textit{t}}}}\colorbox[HTML]{ADCCF6}{\textcolor{black}{\strut{}\textrm{\textit{o}}}} \colorbox[HTML]{ADCCF6}{\textcolor{black}{\strut{}\textrm{\textit{h}}}}\colorbox[HTML]{ADCCF6}{\textcolor{black}{\strut{}\textrm{\textit{a}}}}\colorbox[HTML]{ADCCF6}{\textcolor{black}{\strut{}\textrm{\textit{v}}}}\colorbox[HTML]{ADCCF6}{\textcolor{black}{\strut{}\textrm{\textit{e}}}} \colorbox[HTML]{ADCCF6}{\textcolor{black}{\strut{}\textrm{\textit{l}}}}\colorbox[HTML]{ADCCF6}{\textcolor{black}{\strut{}\textrm{\textit{o}}}}\colorbox[HTML]{ADCCF6}{\textcolor{black}{\strut{}\textrm{\textit{v}}}}\colorbox[HTML]{ADCCF6}{\textcolor{black}{\strut{}\textrm{\textit{e}}}}\colorbox[HTML]{ADCCF6}{\textcolor{black}{\strut{}\textrm{\textit{d}}}} \colorbox[HTML]{FFFFFF}{\textcolor{black}{\strut{}\textrm{\textit{a}}}}\colorbox[HTML]{FFFFFF}{\textcolor{black}{\strut{}\textrm{\textit{n}}}}\colorbox[HTML]{FFFFFF}{\textcolor{black}{\strut{}\textrm{\textit{d}}}} \colorbox[HTML]{ADCCF6}{\textcolor{black}{\strut{}\textrm{\textit{l}}}}\colorbox[HTML]{ADCCF6}{\textcolor{black}{\strut{}\textrm{\textit{o}}}}\colorbox[HTML]{ADCCF6}{\textcolor{black}{\strut{}\textrm{\textit{s}}}}\colorbox[HTML]{ADCCF6}{\textcolor{black}{\strut{}\textrm{\textit{t}}}}\\
    & \colorbox[HTML]{ADCCF6}{\textcolor{black}{\strut{}\textrm{\textit{T}}}}\colorbox[HTML]{ADCCF6}{\textcolor{black}{\strut{}\textrm{\textit{h}}}}\colorbox[HTML]{FFFFFF}{\textcolor{black}{\strut{}\textrm{\textit{a}}}}\colorbox[HTML]{FFFFFF}{\textcolor{black}{\strut{}\textrm{\textit{n}}}} \colorbox[HTML]{ADCCF6}{\textcolor{black}{\strut{}\textrm{\textit{n}}}}\colorbox[HTML]{ADCCF6}{\textcolor{black}{\strut{}\textrm{\textit{e}}}}\colorbox[HTML]{ADCCF6}{\textcolor{black}{\strut{}\textrm{\textit{v}}}}\colorbox[HTML]{ADCCF6}{\textcolor{black}{\strut{}\textrm{\textit{e}}}}\colorbox[HTML]{ADCCF6}{\textcolor{black}{\strut{}\textrm{\textit{r}}}} \colorbox[HTML]{ADCCF6}{\textcolor{black}{\strut{}\textrm{\textit{t}}}}\colorbox[HTML]{ADCCF6}{\textcolor{black}{\strut{}\textrm{\textit{o}}}} \colorbox[HTML]{ADCCF6}{\textcolor{black}{\strut{}\textrm{\textit{h}}}}\colorbox[HTML]{ADCCF6}{\textcolor{black}{\strut{}\textrm{\textit{a}}}}\colorbox[HTML]{ADCCF6}{\textcolor{black}{\strut{}\textrm{\textit{v}}}}\colorbox[HTML]{ADCCF6}{\textcolor{black}{\strut{}\textrm{\textit{e}}}} \colorbox[HTML]{79ABE2}{\textcolor{white}{\strut{}\textrm{\textit{l}}}}\colorbox[HTML]{79ABE2}{\textcolor{white}{\strut{}\textrm{\textit{o}}}}\colorbox[HTML]{79ABE2}{\textcolor{white}{\strut{}\textrm{\textit{v}}}}\colorbox[HTML]{79ABE2}{\textcolor{white}{\strut{}\textrm{\textit{e}}}}\colorbox[HTML]{79ABE2}{\textcolor{white}{\strut{}\textrm{\textit{d}}}} \colorbox[HTML]{00366C}{\textcolor{white}{\strut{}\textrm{\textit{a}}}}\colorbox[HTML]{00366C}{\textcolor{white}{\strut{}\textrm{\textit{t}}}} \colorbox[HTML]{FFFFFF}{\textcolor{black}{\strut{}\textrm{\textit{\textbf{a}}}}}\colorbox[HTML]{FFFFFF}{\textcolor{black}{\strut{}\textrm{\textit{\textbf{l}}}}}\colorbox[HTML]{FFFFFF}{\textcolor{black}{\strut{}\textrm{\textit{\textbf{l}}}}}\colorbox[HTML]{FFFFFF}{\textcolor{black}{\strut{}\textrm{\textit{\textbf{.}}}}}
  \end{tabular}
}%
\global\let\stylesize\undefined%
\global\let\versesize\undefined%

  \caption{A famous stanza by \citet{alfred1850memoriam} with visualized
  attention from \bygpt (top) and \gpt (bottom) when generating the last
  syllable.}%
  \label{fig:visualization}
\end{figure}

To introspect the decision-making processes of our models, we visualize their
token-level attributions when generating a quatrain. Token-level attributions
explain to which degree each token in the input is involved in determining the
next output token of a model, allowing us to reason about what a model has
learned. To this end, \citet{ferrando-etal-2022-mixing} decompose the attention
blocks of transformers into a sum of vectors and define a new measure for
visualizing token-to-token interactions based on the distance of each vector to
the output~\citep{kobayashi-etal-2021-incorporating}. We apply this measure on
generative language models and visualize token-level attributions for \bygpt
and \gpt when generating the last syllable in a quatrain. Since we have
observed reoccurring trends, we use a single visualization in
Figure~\ref{fig:visualization} as a leading example.
We provide additional examples in German in Appendix~\ref{sec:tok-attr-de}.

We can see that \bygpt puts a big emphasis on the current verse, as well as the
styles it was conditioned on. Further, possibly in response to the ABBA rhyme
scheme, it also heavily stresses the ending of the first verse. Since the model
also places a moderate amount of attention on the last consonants in verse
three, it also seems to be aware of which sounds it should \emph{not} generate
in order maintain the rhyme scheme. Interestingly, it heavily emphasizes the
letter \emph{v} in the last two verses. We assume that this corresponds to what
\bygpt understands by alliteration, in which case it would not have understood
well at which position in a word the same sounds must occur.

Unlike \bygpt, \gpt does not put any visible emphasis on input style tokens,
which suggests that it does not understand how to handle them very well.
Nevertheless it stresses similar aspects to \bygpt, although, due to the
subword vocabulary, at a different level of granularity.

\subsection{Comparison with \chatgpt}\label{sec:chatgpt}
\chatgpt~\citep{schulman2022chatgpt} is a conversational large language model
which specializes in dialogue. It has attracted attention for its detailed and
expressive answers, raising the question of how well it performs in generating
poetry. In a small-scale study, we thus ask \chatgpt to generate quatrains with
various rhyme schemes (AABB, ABAB, ABBA, and ABCB) using its web
interface,\footnote{After experimenting with prompt engineering and in-context
learning, we settled on a straightforward template: \enquote{Generate a
quatrain with \texttt{<pattern>} rhyme scheme.}} and similarly generate poems
using \bygpt. We then construct random pairs of quatrains of each model and
want to find out which poem adheres better to rhyme constraints. Since we know
the quatrains of \chatgpt beforehand, we use our rhyme scorer of
\S\ref{sec:auto-eval} for unbiased scoring. Only in 15\% of cases does our
scorer prefer poems of \chatgpt over \bygpt. Manual investigation showed that
\chatgpt tends to generate rhymes at arbitrary positions, rather than adhering
to specified rhyme schemes, even when giving examples in the prompt. Our
verdict is that \chatgpt is a viable approach for poetry generation but not
\emph{style-conditioned} poetry generation.

\section{Related Work}\label{sec:related} 
As indicated in \S\ref{sec:introduction}, competing poetry generation
systems usually consist of model pipelines and/or inject prior knowledge.
\citet{zhang-lapata-2014-chinese}, for example, propose a system for modeling
quatrains consisting of three components: one model encodes previous verses, a
second one reduces them to a single context vector, and a third one generates
verses, one at a time. During decoding, phrases which violate style constraints
are discarded.
Similarly, \textsc{Deep-speare}~\citep{lau-etal-2018-deep} consists of a
language model that generates a set of sample verses in reverse order, a model
that reinitiates sampling as long as rhyme constraints are not met, and a
final model that ranks the samples according to how well they adhere to iambic
pentameter.
\citet{van-de-cruys-2020-automatic} also induce prior knowledge into a
generic language model, but they do it by modifying output probability
distributions directly.
\citet{jhamtani-etal-2019-learning} put their focus on actually \emph{learning}
rhyme and train a sonnet and limerick generator through adversarial training.
The model is hierarchical, i.e., it first generates a sequence of line endings
which are subsequently completed in reverse. While the model manages to learn
the meaning of rhyme to an extent, the authors still filter outputs using
pronunciation dictionaries.
More in line with our research, \citet{hopkins-kiela-2017-automatically} train
a model on the phonetic representation of poetry using the International
Phonetic Alphabet (IPA) as a character-level vocabulary. During inference, a
second model translates sounds back to human-readable text. Although promising,
the model did not generalize well, and an additional model enforces rhythmic
constraints in their final approach.
\citet{ormazabal-2022-poelm} target \emph{unsupervised} poetry generation by
training a system on prosaic text only and conditioning it on structural
information such as line endings and syllable counts. During inference, the
system can generate poetry when conditioned on rhyming line endings and metric
syllable counts. Nonetheless, since this information must be crafted manually,
it is still supervised in a slightly different sense. In contrast, our model is
supervised during training as it requires poetry to learn from, but
unsupervised during inference as it is able to independently incorporate poetic
elements.

\section{Conclusion}\label{sec:conclusion}
In this work, we implement end-to-end style-conditioned poetry generation
systems for quatrains in English and German. Unlike other work, our
systems are able to generate poetry without the need for human supervision,
except for the use of poetic training data.
In particular, we present \bygpt, a novel token-free decoder-only language
model, and show that fine-tuning it on a custom poetry corpus
outperforms other models, such as \gpt, \tfive[m], and \tfive[By], on average,
while also performing favorably against human poets in our constrained setting.
Our key findings are that (i) tokenization algorithms matter, i.e., token-free
language models generally perform better at generating character-level styles
than subword-level transformers, and (ii) large datasets are crucial for
successful training.
We further show that bigger models do not necessarily perform better and that
decoder-only architectures work best, i.e., we can discard the encoder of
\tfive[By] (75\% of parameters) while still
improving downstream performance.
We also demonstrate that token-free transformers perform competitively on tasks
not tied to character-level styles, and are less susceptible to memorization of
the training dataset. In addition, we conduct a visual analysis of token-level
attributions during quatrain generation that is consistent with human
perception of styles.

In future work, we want to to extend our system to other poetic forms such as
sonnets, limericks, or villanelles.

\section{Limitations}\label{sec:limitations} 
\begin{figure}
  \centering
  \includegraphics[width=\textwidth]{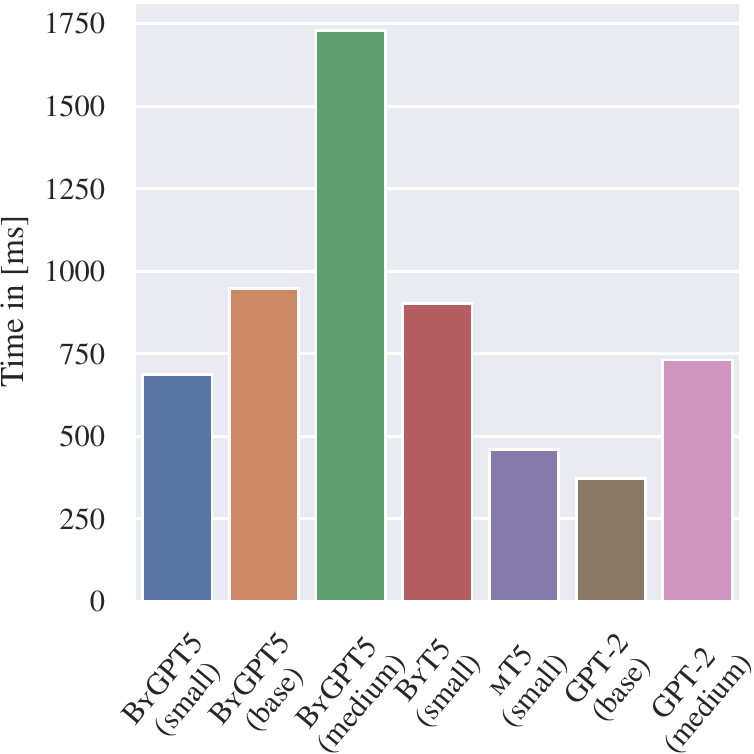}
  \caption{Inference times for generating a single quatrain (with 177
  characters) on an A6000 GPU.}%
  \label{fig:benchmark}
\end{figure}

A well-known shortcoming of transformers is the computational complexity in
self-attention layers~\citep{vaswani2017attention}. Since the number of
required calculations grows quadratically with the length of the input,
transformers become prohibitively slow on very long sequences. An unfortunate
side effect of processing inputs at the character-level is that internal
sequences become much longer, so token-free transformers run into these
efficiency problems much earlier than subword-based models.
Figure~\ref{fig:benchmark} illustrates this problem by contrasting the runtime
of all poetry generation systems when generating a single quatrain.
Even \bygpt (small), the smallest model in terms of number of parameters (cf.
Table~\ref{tab:models}) and the fastest token-free transformer, is only
marginally faster than \gpt (medium), which is almost five times larger.
\citet{tay2022charformer} propose a solution to this problem for transformer
encoder blocks by applying a neural pooling operation over input embeddings
before feeding them into the model, which could be extended to decoder blocks
in future work. Alternatively, \citet{libovicky-etal-2022-dont} propose a
two-stage decoding architecture in which the transformer decoder operates on
character blocks that an additional LSTM model~\citep{hochreiter1997lstm}
decodes into individual characters.

Another shortcoming is that our poetry generation systems can only generate a
single poetic form, i.e., quatrains. In general, poetry is a very diverse form
of language and stanzas can be of arbitrary length, so this is a serious
limitation. In future work, we thus plan to extend our implementation of
style-conditioning to variable length poems. In particular, one could encode a
rhyme scheme not as a single special token, but as an arbitrary series of
letters indicating which verses rhyme with each other. Alternatively, our
current systems could be used to generate longer stanzas through a sliding
window approach, i.e., generating one verse at a time with the last three verse
as context.

Further, our human evaluation has limitations due to its relatively small
scope. We only have a limited number of annotators and only consider a subset
of all style combinations. Nevertheless, we have achieved moderately high to
high agreement on all tasks, and we have an additional human evaluation of
German poetry in Appendix~\ref{sec:demographics}, which points to the same
conclusion.

Lastly, \quatrain is limited in that it consists of pseudo-quatrains, which
are not real quatrains and often have missing contexts. Nonetheless, as can be
seen in Appendix~\ref{sec:example-quatrains}, models trained on \quatrain are
still able to generate meaningful poetry. In future work, we plan to improve
the quality of our dataset by obtaining real quatrains from additional sources
such as the Eighteenth-Century Poetry Archive~\citep{huber2022ecpa}.

\section*{Acknowledgments}
We thank all reviewers for their valuable feedback, hard work, and time. We
also thank all annotators, without whom this work would not have been possible.
The last author was supported by DFG grant EG 375/5--1. We further gratefully
thank the BMBF for its support via the grant Metrics4NLG\@.

\bibliographystyle{acl_natbib}
\bibliography{bibliography}

\clearpage
\appendix
\section{Additional poetry corpus statistics}\label{sec:corpora-statistics} 
The poetry corpora we collect are
English Project Gutenberg (\epg) and
Deutsches Lyrik Korpus (\dlk)~\citep{haider-2021-metrical}
for unlabeled poetry, and
\prosodic\footnote{\url{https://github.com/quadrismegistus/prosodic}},
Chicago Rhyme Corpus~(\chicago)\footnote{\url{https://github.com/sravanareddy/rhymedata}},
For-better-for-verse~(\forb)~\citep{tucker-2022-fbfv},
German Rhyme Corpus (\grc)~\citep{haider-kuhn-2018-supervised},
as well as \epgsf and \antik~\citep{haider-2021-metrical}
for labeled poetry. We map meters which appear less than 25 times in our
labeled corpora to the special label \emph{other}. The final list of meters we
consider can be found in Table~\ref{tab:meters}.
\begin{table}
  \centering
  \begin{tabular}{ll}
    \toprule
    \textbf{Meter} & \textbf{Symbol}\\
    \midrule
    iambus      & \metricsymbols*{u _}\\
    trochee     & \metricsymbols*{_ u}\\
    amphibrach  & \metricsymbols*{u _ u}\\
    anapest    & \metricsymbols*{u u _}\\
    dactyl      & \metricsymbols*{_ u u}\\
    alexandrine & \metricsymbols*{u _ u _ u _ || u _ u _ u _}\\
    \bottomrule
  \end{tabular}
  \caption{Meters in our dataset we consider for our experiments. An
  alexandrine consists of iambic feet with a caesura after the sixth syllable.}%
  \label{tab:meters}
\end{table}

The performance of the meter and rhyme classifiers we train can be seen in
Table~\ref{tab:classification}. For each style, we perform a 90/5/5
train-valid-test split and fine-tune a range of encoder-only transformers with
classification heads jointly on both languages, as this improves
performance~\citep{delaRosa2021,haider-kuhn-2018-supervised}. We test
subword-level \bert[m] and \xlm, as well as character-level
\canine~\citep{clark-etal-2022-canine}. Since character-level \canine
outperforms both \bert[m] and \xlm on both tasks, we use it as our final
classifier.
\begin{table}
  \begin{tabular}{lrr}
    \toprule
    \textbf{Model} & \textbf{Rhyme} & \textbf{Meter}\\
    \midrule
    \canine  & \hl{98.05} & \hl{58.49}\\
    \xlm     & 97.22 & 54.65\\
    \bert[m] & 97.17 & 49.01\\
    \bottomrule
  \end{tabular}
  \caption{F1-Score on classifying rhyme and meter.}%
  \label{tab:classification}
\end{table}

During automatic labeling, when the rhyme scheme cannot be clearly determined
(e.g., according to the classifier the first verse rhymes with the second, the
second with the third but the first and the third do not rhyme) or no dominant
meter exists, we discard the quatrain. Frequencies of automatic labels inside
\quatrain can be seen in Table~\ref{tab:quatrain-stats}.
\begin{table*}
    \begin{tabular}{llrlrlr}
      \toprule
      & \multicolumn{2}{c}{\bf Rhyme} & \multicolumn{2}{c}{\bf Meter} & \multicolumn{2}{c}{\bf Alliteration}\\
      \cmidrule(r){2-3}\cmidrule(lr){4-5}\cmidrule(l){6-7}
      \textbf{Language} & label & freq. & label & freq. & label & freq.\\
      \midrule
      \multirow{3}{*}{German}
      & ABCD & 19.73\% & iambus      & 61.66\% & low    & 50.95\%\\
      & ABAB & 15.60\% & alexandrine & 18.06\% & medium & 39.23\%\\
      & AABB & 13.07\% & trochee     & 17.05\% & high   & 9.81\%\\
      \midrule
      \multirow{3}{*}{English}
      & AABB & 19.17\% & iambus      & 83.96\% & medium & 54.92\%\\
      & ABCD & 16.40\% & anapest    &  7.61\% & low    & 28.94\%\\
      & ABBC & 13.07\% & trochee     &  4.13\% & high   & 16.14\%\\
      \bottomrule
    \end{tabular}
    \caption{Distribution of alliteration levels, as well as most frequent
    meters and rhyme schemes in \quatrain.}%
    \label{tab:quatrain-stats}
\end{table*}
By limiting \quatrain to quatrains, we not only reduce the burden on poetry
generation systems by considering only a single poetic form, but also ease the
labeling process. As the length of poems increases, the number of verse pairs
that have to be classified for rhyme grows super-exponentially, which quickly
becomes intractable.

The eight emotions in \poemo we train our classifier on are \emph{beauty /
joy}, \emph{sadness}, \emph{uneasiness}, \emph{vitality}, \emph{awe / sublime},
\emph{suspense}, \emph{humor}, and \emph{annoyance}. Since an additional
emotion, \emph{nostalgia}, almost never occurs, we follow
\citet{haider-etal-2020-po} and omit it from our experiments.

\section{Annotator Demographics}\label{sec:demographics}
Our human annotators are fluent in English at a C1 or higher level according to
the Common European Framework of Reference for Languages (CEFR). The annotators
for rhyme and human likeness are one male faculty member, two male PhD
students, one female undergraduate student, and three female volunteers from
other departments, amounting to seven distinct annotators who are all
proficient in English but may have limited knowledge of poetry. To get four
sets of annotations per style, we have one PhD student annotate both styles.
For meter, we hire a professional female teacher who is specialized in English
and music.

Since none of our annotators speak English as a native language, we have one
PhD student and one faculty member conduct a small-scale comparative study in
their native language, German, annotating 30 BWS tuples for rhyme and human
likeness. The results in Table~\ref{tab:german-bws} confirm the trends we saw
in human-evaluation in \S\ref{sec:human-eval}. In terms of rhyme, humans
perform best, followed by \bygpt, \tfive[By], \tfive[m], and finally \gpt. On
human likeness \bygpt is outperformed by humans but performs similar to \gpt.
The encoder-decoder models \tfive[m] and \tfive[By] perform the worst, likely
for similar reasons outlined in \S\ref{sec:human-eval}.

\section{German Token-level Attributions}\label{sec:tok-attr-de}
Figure~\ref{fig:visualization-de} shows token-level attribution scores for
German quatrains. By and large, we observe the same trends as in
\S\ref{sec:tok-attr} on English, i.e., \gpt places less attention on
style and the emphasized parts of the text are less granular.
\begin{figure*}
\begin{floatrow}
\ttabbox{%
  \begin{tabular}{lrr}
    \toprule
    \textbf{Model} & \textbf{Rhyme} & \textbf{Human Likeness}\\
    \midrule
    Human          & \first{0.84}   & \first{0.89}\\
    \bygpt         & \second{0.57}  & \second{0.55}\\
    \gpt           & 0.35           & \second{0.55}\\
    \tfive[By]     & 0.45           & 0.19\\
    \tfive[m]      & 0.28           & 0.32\\
    \bottomrule
  \end{tabular}%
}{%
  \caption{Min-max normalized and averaged BWS scores annotated by two native
  German speakers. The SHR is $\rho = 0.83$ for rhyme and $\rho = 0.61$ for
  human likeness.}%
  \label{tab:german-bws}%
}
\ffigbox{%

\ifx\versesize\undefined\def\versesize{\normalsize}\fi%
\ifx\stylesize\undefined\def\stylesize{\small}\fi%
{\versesize\fboxsep0pt{}
  \begin{tabular}{cl}
    \multirow{4}{*}{\stylesize\makecell{\colorbox[HTML]{ADCCF6}{\textcolor{black}{\strut{}{\texttt{<ABBA>}}}}\\\colorbox[HTML]{ADCCF6}{\textcolor{black}{\strut{}{\texttt{<trochee>}}}}\\\colorbox[HTML]{FFFFFF}{\textcolor{black}{\strut{}{\texttt{<medium>}}}}}}
    & \colorbox[HTML]{ADCCF6}{\textcolor{black}{\strut{}\textrm{\textit{F}}}}\colorbox[HTML]{FFFFFF}{\textcolor{black}{\strut{}\textrm{\textit{r}}}}\colorbox[HTML]{FFFFFF}{\textcolor{black}{\strut{}\textrm{\textit{ü}}}}\colorbox[HTML]{FFFFFF}{\textcolor{black}{\strut{}\textrm{\textit{h}}}}\colorbox[HTML]{FFFFFF}{\textcolor{black}{\strut{}\textrm{\textit{l}}}}\colorbox[HTML]{FFFFFF}{\textcolor{black}{\strut{}\textrm{\textit{i}}}}\colorbox[HTML]{FFFFFF}{\textcolor{black}{\strut{}\textrm{\textit{n}}}}\colorbox[HTML]{FFFFFF}{\textcolor{black}{\strut{}\textrm{\textit{g}}}} \colorbox[HTML]{FFFFFF}{\textcolor{black}{\strut{}\textrm{\textit{l}}}}\colorbox[HTML]{ADCCF6}{\textcolor{black}{\strut{}\textrm{\textit{ä}}}}\colorbox[HTML]{ADCCF6}{\textcolor{black}{\strut{}\textrm{\textit{ß}}}}\colorbox[HTML]{FFFFFF}{\textcolor{black}{\strut{}\textrm{\textit{t}}}} \colorbox[HTML]{FFFFFF}{\textcolor{black}{\strut{}\textrm{\textit{s}}}}\colorbox[HTML]{FFFFFF}{\textcolor{black}{\strut{}\textrm{\textit{e}}}}\colorbox[HTML]{FFFFFF}{\textcolor{black}{\strut{}\textrm{\textit{i}}}}\colorbox[HTML]{FFFFFF}{\textcolor{black}{\strut{}\textrm{\textit{n}}}} \colorbox[HTML]{ADCCF6}{\textcolor{black}{\strut{}\textrm{\textit{b}}}}\colorbox[HTML]{ADCCF6}{\textcolor{black}{\strut{}\textrm{\textit{l}}}}\colorbox[HTML]{ADCCF6}{\textcolor{black}{\strut{}\textrm{\textit{a}}}}\colorbox[HTML]{ADCCF6}{\textcolor{black}{\strut{}\textrm{\textit{u}}}}\colorbox[HTML]{ADCCF6}{\textcolor{black}{\strut{}\textrm{\textit{e}}}}\colorbox[HTML]{ADCCF6}{\textcolor{black}{\strut{}\textrm{\textit{s}}}} \colorbox[HTML]{00366C}{\textcolor{white}{\strut{}\textrm{\textit{B}}}}\colorbox[HTML]{79ABE2}{\textcolor{white}{\strut{}\textrm{\textit{a}}}}\colorbox[HTML]{79ABE2}{\textcolor{white}{\strut{}\textrm{\textit{n}}}}\colorbox[HTML]{2C86CA}{\textcolor{white}{\strut{}\textrm{\textit{d}}}}\\
    & \colorbox[HTML]{FFFFFF}{\textcolor{black}{\strut{}\textrm{\textit{W}}}}\colorbox[HTML]{FFFFFF}{\textcolor{black}{\strut{}\textrm{\textit{i}}}}\colorbox[HTML]{FFFFFF}{\textcolor{black}{\strut{}\textrm{\textit{e}}}}\colorbox[HTML]{FFFFFF}{\textcolor{black}{\strut{}\textrm{\textit{d}}}}\colorbox[HTML]{FFFFFF}{\textcolor{black}{\strut{}\textrm{\textit{e}}}}\colorbox[HTML]{FFFFFF}{\textcolor{black}{\strut{}\textrm{\textit{r}}}} \colorbox[HTML]{FFFFFF}{\textcolor{black}{\strut{}\textrm{\textit{f}}}}\colorbox[HTML]{FFFFFF}{\textcolor{black}{\strut{}\textrm{\textit{l}}}}\colorbox[HTML]{FFFFFF}{\textcolor{black}{\strut{}\textrm{\textit{a}}}}\colorbox[HTML]{FFFFFF}{\textcolor{black}{\strut{}\textrm{\textit{t}}}}\colorbox[HTML]{FFFFFF}{\textcolor{black}{\strut{}\textrm{\textit{t}}}}\colorbox[HTML]{FFFFFF}{\textcolor{black}{\strut{}\textrm{\textit{e}}}}\colorbox[HTML]{FFFFFF}{\textcolor{black}{\strut{}\textrm{\textit{r}}}}\colorbox[HTML]{FFFFFF}{\textcolor{black}{\strut{}\textrm{\textit{n}}}} \colorbox[HTML]{FFFFFF}{\textcolor{black}{\strut{}\textrm{\textit{d}}}}\colorbox[HTML]{FFFFFF}{\textcolor{black}{\strut{}\textrm{\textit{u}}}}\colorbox[HTML]{FFFFFF}{\textcolor{black}{\strut{}\textrm{\textit{r}}}}\colorbox[HTML]{FFFFFF}{\textcolor{black}{\strut{}\textrm{\textit{c}}}}\colorbox[HTML]{FFFFFF}{\textcolor{black}{\strut{}\textrm{\textit{h}}}} \colorbox[HTML]{FFFFFF}{\textcolor{black}{\strut{}\textrm{\textit{d}}}}\colorbox[HTML]{FFFFFF}{\textcolor{black}{\strut{}\textrm{\textit{i}}}}\colorbox[HTML]{FFFFFF}{\textcolor{black}{\strut{}\textrm{\textit{e}}}} \colorbox[HTML]{FFFFFF}{\textcolor{black}{\strut{}\textrm{\textit{L}}}}\colorbox[HTML]{FFFFFF}{\textcolor{black}{\strut{}\textrm{\textit{ü}}}}\colorbox[HTML]{FFFFFF}{\textcolor{black}{\strut{}\textrm{\textit{f}}}}\colorbox[HTML]{FFFFFF}{\textcolor{black}{\strut{}\textrm{\textit{t}}}}\colorbox[HTML]{FFFFFF}{\textcolor{black}{\strut{}\textrm{\textit{e}}}}\colorbox[HTML]{ADCCF6}{\textcolor{black}{\strut{}\textrm{\textit{;}}}}\\
    & \colorbox[HTML]{FFFFFF}{\textcolor{black}{\strut{}\textrm{\textit{S}}}}\colorbox[HTML]{FFFFFF}{\textcolor{black}{\strut{}\textrm{\textit{ü}}}}\colorbox[HTML]{FFFFFF}{\textcolor{black}{\strut{}\textrm{\textit{ß}}}}\colorbox[HTML]{FFFFFF}{\textcolor{black}{\strut{}\textrm{\textit{e}}}}\colorbox[HTML]{ADCCF6}{\textcolor{black}{\strut{}\textrm{\textit{,}}}} \colorbox[HTML]{FFFFFF}{\textcolor{black}{\strut{}\textrm{\textit{w}}}}\colorbox[HTML]{FFFFFF}{\textcolor{black}{\strut{}\textrm{\textit{o}}}}\colorbox[HTML]{FFFFFF}{\textcolor{black}{\strut{}\textrm{\textit{h}}}}\colorbox[HTML]{FFFFFF}{\textcolor{black}{\strut{}\textrm{\textit{l}}}}\colorbox[HTML]{FFFFFF}{\textcolor{black}{\strut{}\textrm{\textit{b}}}}\colorbox[HTML]{FFFFFF}{\textcolor{black}{\strut{}\textrm{\textit{e}}}}\colorbox[HTML]{FFFFFF}{\textcolor{black}{\strut{}\textrm{\textit{k}}}}\colorbox[HTML]{FFFFFF}{\textcolor{black}{\strut{}\textrm{\textit{a}}}}\colorbox[HTML]{FFFFFF}{\textcolor{black}{\strut{}\textrm{\textit{n}}}}\colorbox[HTML]{FFFFFF}{\textcolor{black}{\strut{}\textrm{\textit{n}}}}\colorbox[HTML]{FFFFFF}{\textcolor{black}{\strut{}\textrm{\textit{t}}}}\colorbox[HTML]{FFFFFF}{\textcolor{black}{\strut{}\textrm{\textit{e}}}} \colorbox[HTML]{ADCCF6}{\textcolor{black}{\strut{}\textrm{\textit{D}}}}\colorbox[HTML]{ADCCF6}{\textcolor{black}{\strut{}\textrm{\textit{ü}}}}\colorbox[HTML]{ADCCF6}{\textcolor{black}{\strut{}\textrm{\textit{f}}}}\colorbox[HTML]{ADCCF6}{\textcolor{black}{\strut{}\textrm{\textit{t}}}}\colorbox[HTML]{ADCCF6}{\textcolor{black}{\strut{}\textrm{\textit{e}}}}\\
    & \colorbox[HTML]{79ABE2}{\textcolor{white}{\strut{}\textrm{\textit{S}}}}\colorbox[HTML]{ADCCF6}{\textcolor{black}{\strut{}\textrm{\textit{t}}}}\colorbox[HTML]{ADCCF6}{\textcolor{black}{\strut{}\textrm{\textit{r}}}}\colorbox[HTML]{ADCCF6}{\textcolor{black}{\strut{}\textrm{\textit{e}}}}\colorbox[HTML]{79ABE2}{\textcolor{white}{\strut{}\textrm{\textit{i}}}}\colorbox[HTML]{005D9A}{\textcolor{white}{\strut{}\textrm{\textit{f}}}}\colorbox[HTML]{ADCCF6}{\textcolor{black}{\strut{}\textrm{\textit{e}}}}\colorbox[HTML]{FFFFFF}{\textcolor{black}{\strut{}\textrm{\textit{n}}}} \colorbox[HTML]{ADCCF6}{\textcolor{black}{\strut{}\textrm{\textit{a}}}}\colorbox[HTML]{ADCCF6}{\textcolor{black}{\strut{}\textrm{\textit{h}}}}\colorbox[HTML]{ADCCF6}{\textcolor{black}{\strut{}\textrm{\textit{n}}}}\colorbox[HTML]{ADCCF6}{\textcolor{black}{\strut{}\textrm{\textit{u}}}}\colorbox[HTML]{ADCCF6}{\textcolor{black}{\strut{}\textrm{\textit{n}}}}\colorbox[HTML]{ADCCF6}{\textcolor{black}{\strut{}\textrm{\textit{g}}}}\colorbox[HTML]{ADCCF6}{\textcolor{black}{\strut{}\textrm{\textit{s}}}}\colorbox[HTML]{005D9A}{\textcolor{white}{\strut{}\textrm{\textit{v}}}}\colorbox[HTML]{ADCCF6}{\textcolor{black}{\strut{}\textrm{\textit{o}}}}\colorbox[HTML]{79ABE2}{\textcolor{white}{\strut{}\textrm{\textit{l}}}}\colorbox[HTML]{79ABE2}{\textcolor{white}{\strut{}\textrm{\textit{l}}}} \colorbox[HTML]{79ABE2}{\textcolor{white}{\strut{}\textrm{\textit{d}}}}\colorbox[HTML]{79ABE2}{\textcolor{white}{\strut{}\textrm{\textit{a}}}}\colorbox[HTML]{79ABE2}{\textcolor{white}{\strut{}\textrm{\textit{s}}}} \colorbox[HTML]{FFFFFF}{\textcolor{black}{\strut{}\textrm{\textit{\textbf{L}}}}}\colorbox[HTML]{FFFFFF}{\textcolor{black}{\strut{}\textrm{\textit{\textbf{a}}}}}\colorbox[HTML]{FFFFFF}{\textcolor{black}{\strut{}\textrm{\textit{\textbf{n}}}}}\colorbox[HTML]{FFFFFF}{\textcolor{black}{\strut{}\textrm{\textit{\textbf{d}}}}}\colorbox[HTML]{FFFFFF}{\textcolor{black}{\strut{}\textrm{\textit{\textbf{.}}}}}
  \end{tabular}
}%
\global\let\stylesize\undefined%
\global\let\versesize\undefined%

  \bigskip

\ifx\versesize\undefined\def\versesize{\normalsize}\fi%
\ifx\stylesize\undefined\def\stylesize{\small}\fi%
{\versesize\fboxsep0pt{}
  \begin{tabular}{cl}
    \multirow{4}{*}{\stylesize\makecell{\colorbox[HTML]{ADCCF6}{\textcolor{black}{\strut{}{\texttt{<ABBA>}}}}\\\colorbox[HTML]{FFFFFF}{\textcolor{black}{\strut{}{\texttt{<trochee>}}}}\\\colorbox[HTML]{FFFFFF}{\textcolor{black}{\strut{}{\texttt{<medium>}}}}}}
    & \colorbox[HTML]{FFFFFF}{\textcolor{black}{\strut{}\textrm{\textit{F}}}}\colorbox[HTML]{FFFFFF}{\textcolor{black}{\strut{}\textrm{\textit{r}}}}\colorbox[HTML]{FFFFFF}{\textcolor{black}{\strut{}\textrm{\textit{ü}}}}\colorbox[HTML]{ADCCF6}{\textcolor{black}{\strut{}\textrm{\textit{h}}}}\colorbox[HTML]{ADCCF6}{\textcolor{black}{\strut{}\textrm{\textit{l}}}}\colorbox[HTML]{ADCCF6}{\textcolor{black}{\strut{}\textrm{\textit{i}}}}\colorbox[HTML]{ADCCF6}{\textcolor{black}{\strut{}\textrm{\textit{n}}}}\colorbox[HTML]{ADCCF6}{\textcolor{black}{\strut{}\textrm{\textit{g}}}} \colorbox[HTML]{FFFFFF}{\textcolor{black}{\strut{}\textrm{\textit{l}}}}\colorbox[HTML]{FFFFFF}{\textcolor{black}{\strut{}\textrm{\textit{ä}}}}\colorbox[HTML]{FFFFFF}{\textcolor{black}{\strut{}\textrm{\textit{ß}}}}\colorbox[HTML]{FFFFFF}{\textcolor{black}{\strut{}\textrm{\textit{t}}}} \colorbox[HTML]{FFFFFF}{\textcolor{black}{\strut{}\textrm{\textit{s}}}}\colorbox[HTML]{FFFFFF}{\textcolor{black}{\strut{}\textrm{\textit{e}}}}\colorbox[HTML]{FFFFFF}{\textcolor{black}{\strut{}\textrm{\textit{i}}}}\colorbox[HTML]{FFFFFF}{\textcolor{black}{\strut{}\textrm{\textit{n}}}} \colorbox[HTML]{ADCCF6}{\textcolor{black}{\strut{}\textrm{\textit{b}}}}\colorbox[HTML]{ADCCF6}{\textcolor{black}{\strut{}\textrm{\textit{l}}}}\colorbox[HTML]{ADCCF6}{\textcolor{black}{\strut{}\textrm{\textit{a}}}}\colorbox[HTML]{ADCCF6}{\textcolor{black}{\strut{}\textrm{\textit{u}}}}\colorbox[HTML]{ADCCF6}{\textcolor{black}{\strut{}\textrm{\textit{e}}}}\colorbox[HTML]{ADCCF6}{\textcolor{black}{\strut{}\textrm{\textit{s}}}} \colorbox[HTML]{2C86CA}{\textcolor{white}{\strut{}\textrm{\textit{B}}}}\colorbox[HTML]{2C86CA}{\textcolor{white}{\strut{}\textrm{\textit{a}}}}\colorbox[HTML]{2C86CA}{\textcolor{white}{\strut{}\textrm{\textit{n}}}}\colorbox[HTML]{2C86CA}{\textcolor{white}{\strut{}\textrm{\textit{d}}}}\\
    & \colorbox[HTML]{FFFFFF}{\textcolor{black}{\strut{}\textrm{\textit{W}}}}\colorbox[HTML]{FFFFFF}{\textcolor{black}{\strut{}\textrm{\textit{i}}}}\colorbox[HTML]{FFFFFF}{\textcolor{black}{\strut{}\textrm{\textit{e}}}}\colorbox[HTML]{FFFFFF}{\textcolor{black}{\strut{}\textrm{\textit{d}}}}\colorbox[HTML]{FFFFFF}{\textcolor{black}{\strut{}\textrm{\textit{e}}}}\colorbox[HTML]{FFFFFF}{\textcolor{black}{\strut{}\textrm{\textit{r}}}} \colorbox[HTML]{FFFFFF}{\textcolor{black}{\strut{}\textrm{\textit{f}}}}\colorbox[HTML]{FFFFFF}{\textcolor{black}{\strut{}\textrm{\textit{l}}}}\colorbox[HTML]{FFFFFF}{\textcolor{black}{\strut{}\textrm{\textit{a}}}}\colorbox[HTML]{FFFFFF}{\textcolor{black}{\strut{}\textrm{\textit{t}}}}\colorbox[HTML]{FFFFFF}{\textcolor{black}{\strut{}\textrm{\textit{t}}}}\colorbox[HTML]{FFFFFF}{\textcolor{black}{\strut{}\textrm{\textit{e}}}}\colorbox[HTML]{FFFFFF}{\textcolor{black}{\strut{}\textrm{\textit{r}}}}\colorbox[HTML]{FFFFFF}{\textcolor{black}{\strut{}\textrm{\textit{n}}}} \colorbox[HTML]{FFFFFF}{\textcolor{black}{\strut{}\textrm{\textit{d}}}}\colorbox[HTML]{FFFFFF}{\textcolor{black}{\strut{}\textrm{\textit{u}}}}\colorbox[HTML]{FFFFFF}{\textcolor{black}{\strut{}\textrm{\textit{r}}}}\colorbox[HTML]{FFFFFF}{\textcolor{black}{\strut{}\textrm{\textit{c}}}}\colorbox[HTML]{FFFFFF}{\textcolor{black}{\strut{}\textrm{\textit{h}}}} \colorbox[HTML]{ADCCF6}{\textcolor{black}{\strut{}\textrm{\textit{d}}}}\colorbox[HTML]{ADCCF6}{\textcolor{black}{\strut{}\textrm{\textit{i}}}}\colorbox[HTML]{ADCCF6}{\textcolor{black}{\strut{}\textrm{\textit{e}}}} \colorbox[HTML]{FFFFFF}{\textcolor{black}{\strut{}\textrm{\textit{L}}}}\colorbox[HTML]{FFFFFF}{\textcolor{black}{\strut{}\textrm{\textit{ü}}}}\colorbox[HTML]{FFFFFF}{\textcolor{black}{\strut{}\textrm{\textit{f}}}}\colorbox[HTML]{FFFFFF}{\textcolor{black}{\strut{}\textrm{\textit{t}}}}\colorbox[HTML]{FFFFFF}{\textcolor{black}{\strut{}\textrm{\textit{e}}}}\colorbox[HTML]{FFFFFF}{\textcolor{black}{\strut{}\textrm{\textit{;}}}}\\
    & \colorbox[HTML]{FFFFFF}{\textcolor{black}{\strut{}\textrm{\textit{S}}}}\colorbox[HTML]{FFFFFF}{\textcolor{black}{\strut{}\textrm{\textit{ü}}}}\colorbox[HTML]{FFFFFF}{\textcolor{black}{\strut{}\textrm{\textit{ß}}}}\colorbox[HTML]{FFFFFF}{\textcolor{black}{\strut{}\textrm{\textit{e}}}}\colorbox[HTML]{FFFFFF}{\textcolor{black}{\strut{}\textrm{\textit{,}}}} \colorbox[HTML]{FFFFFF}{\textcolor{black}{\strut{}\textrm{\textit{w}}}}\colorbox[HTML]{FFFFFF}{\textcolor{black}{\strut{}\textrm{\textit{o}}}}\colorbox[HTML]{FFFFFF}{\textcolor{black}{\strut{}\textrm{\textit{h}}}}\colorbox[HTML]{FFFFFF}{\textcolor{black}{\strut{}\textrm{\textit{l}}}}\colorbox[HTML]{ADCCF6}{\textcolor{black}{\strut{}\textrm{\textit{b}}}}\colorbox[HTML]{ADCCF6}{\textcolor{black}{\strut{}\textrm{\textit{e}}}}\colorbox[HTML]{ADCCF6}{\textcolor{black}{\strut{}\textrm{\textit{k}}}}\colorbox[HTML]{ADCCF6}{\textcolor{black}{\strut{}\textrm{\textit{a}}}}\colorbox[HTML]{ADCCF6}{\textcolor{black}{\strut{}\textrm{\textit{n}}}}\colorbox[HTML]{ADCCF6}{\textcolor{black}{\strut{}\textrm{\textit{n}}}}\colorbox[HTML]{ADCCF6}{\textcolor{black}{\strut{}\textrm{\textit{t}}}}\colorbox[HTML]{ADCCF6}{\textcolor{black}{\strut{}\textrm{\textit{e}}}} \colorbox[HTML]{ADCCF6}{\textcolor{black}{\strut{}\textrm{\textit{D}}}}\colorbox[HTML]{ADCCF6}{\textcolor{black}{\strut{}\textrm{\textit{ü}}}}\colorbox[HTML]{ADCCF6}{\textcolor{black}{\strut{}\textrm{\textit{f}}}}\colorbox[HTML]{ADCCF6}{\textcolor{black}{\strut{}\textrm{\textit{t}}}}\colorbox[HTML]{ADCCF6}{\textcolor{black}{\strut{}\textrm{\textit{e}}}}\\
    & \colorbox[HTML]{00366C}{\textcolor{white}{\strut{}\textrm{\textit{S}}}}\colorbox[HTML]{00366C}{\textcolor{white}{\strut{}\textrm{\textit{t}}}}\colorbox[HTML]{00366C}{\textcolor{white}{\strut{}\textrm{\textit{r}}}}\colorbox[HTML]{00366C}{\textcolor{white}{\strut{}\textrm{\textit{e}}}}\colorbox[HTML]{00366C}{\textcolor{white}{\strut{}\textrm{\textit{i}}}}\colorbox[HTML]{00366C}{\textcolor{white}{\strut{}\textrm{\textit{f}}}}\colorbox[HTML]{00366C}{\textcolor{white}{\strut{}\textrm{\textit{e}}}}\colorbox[HTML]{00366C}{\textcolor{white}{\strut{}\textrm{\textit{n}}}} \colorbox[HTML]{ADCCF6}{\textcolor{black}{\strut{}\textrm{\textit{a}}}}\colorbox[HTML]{ADCCF6}{\textcolor{black}{\strut{}\textrm{\textit{h}}}}\colorbox[HTML]{ADCCF6}{\textcolor{black}{\strut{}\textrm{\textit{n}}}}\colorbox[HTML]{ADCCF6}{\textcolor{black}{\strut{}\textrm{\textit{u}}}}\colorbox[HTML]{ADCCF6}{\textcolor{black}{\strut{}\textrm{\textit{n}}}}\colorbox[HTML]{ADCCF6}{\textcolor{black}{\strut{}\textrm{\textit{g}}}}\colorbox[HTML]{ADCCF6}{\textcolor{black}{\strut{}\textrm{\textit{s}}}}\colorbox[HTML]{2C86CA}{\textcolor{white}{\strut{}\textrm{\textit{v}}}}\colorbox[HTML]{2C86CA}{\textcolor{white}{\strut{}\textrm{\textit{o}}}}\colorbox[HTML]{2C86CA}{\textcolor{white}{\strut{}\textrm{\textit{l}}}}\colorbox[HTML]{2C86CA}{\textcolor{white}{\strut{}\textrm{\textit{l}}}} \colorbox[HTML]{00366C}{\textcolor{white}{\strut{}\textrm{\textit{d}}}}\colorbox[HTML]{00366C}{\textcolor{white}{\strut{}\textrm{\textit{a}}}}\colorbox[HTML]{00366C}{\textcolor{white}{\strut{}\textrm{\textit{s}}}} \colorbox[HTML]{FFFFFF}{\textcolor{black}{\strut{}\textrm{\textit{\textbf{L}}}}}\colorbox[HTML]{FFFFFF}{\textcolor{black}{\strut{}\textrm{\textit{\textbf{a}}}}}\colorbox[HTML]{FFFFFF}{\textcolor{black}{\strut{}\textrm{\textit{\textbf{n}}}}}\colorbox[HTML]{FFFFFF}{\textcolor{black}{\strut{}\textrm{\textit{\textbf{d}}}}}\colorbox[HTML]{FFFFFF}{\textcolor{black}{\strut{}\textrm{\textit{\textbf{.}}}}}
  \end{tabular}
}%
\global\let\stylesize\undefined%
\global\let\versesize\undefined%
}{%
  \caption{A well-known stanza by \citet{moerike1832nolten} with visualized
  attention from \bygpt (top) and \gpt (bottom) when generating the last
  syllable.}%
  \label{fig:visualization-de}%
}
\end{floatrow}
\end{figure*}

\section{Example Quatrains}\label{sec:example-quatrains}
In Table~\ref{tab:additional-examples} we list additional example quatrains in
German and English, generated with \bygpt (base).

\begin{table*}
  \begin{tabular}{ll}
    \toprule
    \multicolumn{1}{c}{\textbf{German}} & \multicolumn{1}{c}{\textbf{English}}\\
    \midrule\\[-\ht\strutbox]
    {\stanza{ABAB}
      {\metrics{u _ u _ u _ u}
        {Ein Rei-ter steht am Ha-\A{fen},}}
      {\metrics{u _ u _ u _}
        {Der schaut die Flut nicht \B{an},}}
      {\metrics{u _ u _ u _ u}
        {Er hört die \allit{Sch}if-fer \allit{sch}la-\A{fen}}}
      {\metrics{u _ u _ u _}
          {Im stil-len O-ze-\B{an}.}}} &
    {\stanza{AABB}
      {\metrics{u _ u _ u _ u _ u _ u}
        {With lang-uid \allit{s}mile, the \allit{s}tea-ling \allit{t}ear re-\allit{t}i-\A{res},}}
      {\metrics{u _ u _ u _ u _ u _ u}
        {And the slow fa-ding light on tremb-ling fi-\A{res}!}}
      {\metrics{u _ u _ u _ u _ u _ u}
        {Now she re-cei-ves the gol-den circ-let \B{round},}}
      {\metrics{u _ u _ u _ u _ u _}
        {And fills the \allit{w}o-\allit{v}en cham-bers with a \B{sound};}}}\\[3.5\ht\strutbox]
\midrule\\[-\ht\strutbox]
    {\stanza{ABAB}
     {\metrics{_ u _ u _ u _ u}
       {\allit{Sch}wei-gend \allit{s}tehn die Bur-gen nie-\A{der},}}
     {\metrics{_ u _ u _ u _}
       {Und die Lüf-te sind ver-\B{hallt},}}
     {\metrics{_ u _ u _ u _ u}
       {Und die Trom-meln klin-gen wi-\A{der},}}
     {\metrics{_ u _ u _ u _}
         {Und die Büch-sen knal-len \A{halt}.}}} &
    {\stanza{AABB}
     {\metrics{u _ u _ u _ u _}
       {The first who \allit{l}earned the \allit{l}es-son \A{there}}}
     {\metrics{u _ u _ u _ u _}
       {Had learned to \allit{s}coff and \allit{s}corn to \A{\allit{s}neer},}}
     {\metrics{u _ u _ u _ u _}
       {And \allit{t}hat \allit{t}he lear-ned might have \B{been}}}
     {\metrics{u _ u _ u _ u _}
       {\allit{A} shame-less wo-man \allit{a}nd \allit{a} \B{queen}.}}}\\[3.5\ht\strutbox]
\midrule\\[-\ht\strutbox]
     {\stanza{ABCB}
       {\metrics{u _ u _ u _ u _}
         {Der Greis er-bebt, die Hand er-\A{starrt},}}
       {\metrics{u _ u _ u _ u _ u}
         {\allit{D}ie Kin-\allit{d}er schau-ern vor \allit{d}em Ster-\B{ben};}}
       {\metrics{u _ u _ u _ u _}
         {Die Stim-me bricht, die Thrä-ne \C{fallt},}}
       {\metrics{u _ u _ u _ u _ u}
           {\allit{S}ie \allit{s}ieht ihm nach mit nas-\allit{s}em \allit{B}e-\B{\allit{b}en}.}}} &
     {\stanza{ABAB}
       {\metrics{u _ u _ u _ u _ u _}
         {\allit{Th}en came \allit{th}e la-bour of \allit{th}e day with-\A{in},}}
       {\metrics{u _ u _ u _ u _}
         {The \allit{g}ray be-\allit{g}in-ning of the \B{week},}}
       {\metrics{u _ u _ u _ u _ u _}
         {And down \allit{w}e \allit{w}ent \allit{w}ith hope and ter-ror \A{sin},}}
       {\metrics{u _ u _ u _ u _}
         {\allit{A}nd not \allit{a} word to say and \B{speak}.}}}\\[3.5\ht\strutbox]
\midrule\\[-\ht\strutbox]
      {\stanza{AABB}
        {\metrics{u _ u _ u _ u _ u _ u _}
          {\allit{D}ie Ström-ung wie-\allit{d}er-um \allit{d}urch al-le Glie-\allit{d}er \A{\allit{d}ringt},}}
        {\metrics{u _ u _ u _ u _ u _ u _}
          {Und al-les, was da lebt und wallt und leuch-tet, \A{singt}.}}
        {\metrics{u _ u _ u _ u _ u _ u _ u}
          {Da steigt ein Pal-men-strauch aus dem er-hab-nen Lich-\B{te},}}
        {\metrics{u _ u _ u _ u _ u _ u _ u}
          {\allit{E}r schwimmt \allit{a}uf \allit{e}i-ner Fluth \allit{u}nd sin-get in Ge-dich-\B{te}.}}} &
        {\stanza{AABB}
         {\metrics{_ u _ u _ u _}
           {For the stars are in the \A{sky};}}
         {\metrics{_ u _ u _ u _}
           {And the stars have gone to \B{die}}}
         {\metrics{_ u _ u _ u _}
           {With their songs of joy and \A{fear},}}
         {\metrics{_ u _ u _ u _}
           {With their mu-sic and their \B{cheer}.}}}\\[3.5\ht\strutbox]
    \bottomrule
  \end{tabular}
  \caption{Additional example poems generated with \bygpt (base) in German and English.}%
  \label{tab:additional-examples}
\end{table*}

\end{document}